\documentclass[]{spie}  
\usepackage{amsmath,amsfonts,amssymb}
\usepackage[caption=false]{subfig}
\usepackage{authblk}
\usepackage{blindtext,graphicx}
\usepackage[absolute]{textpos}
\usepackage[colorlinks=true, allcolors=blue]{hyperref}
\usepackage{multicol}

\begin{document}

\begin{textblock}{20}(4,1)
	\normalsize Distribution Statement A: Approved for public release: distribution is unlimited
	\vspace{-22mm}
\end{textblock}

\title{\vspace{9mm}Possibilistic Fuzzy Local Information C-Means with Automated Feature Selection for Seafloor Segmentation}

\author[a]{Joshua Peeples}
\author[a]{Daniel Suen}
\author[a]{Alina Zare}
\author[b]{James Keller}
\affil[a]{Electrical and Computer Engineering, University of Florida,Gainesville, FL, 32611}
\affil[b]{Computer Science and Electrical Engineering, University of Missouri,Columbia, MO, 65211}


\pagestyle{empty} 
\setcounter{page}{301} 

\maketitle

\begin{abstract}
The Possibilistic Fuzzy Local Information C-Means (PFLICM) method is presented as a technique to segment side-look synthetic aperture sonar (SAS) imagery into distinct regions of the sea-floor. In this work, we investigate and present the results of an automated feature selection approach for SAS image segmentation. The chosen features and resulting segmentation from the image will be assessed based on a select quantitative clustering validity criterion and the subset of the features that reach a desired threshold will be used for the segmentation process. 
\end{abstract}

\keywords{Feature selection, SAS, segmentation, clustering, SONAR, cluster validity}

\section{INTRODUCTION}
\label{sec:intro}  

Synthetic aperture sonar (SAS) produces high resolution images of the sea environment \cite{hayes2009synthetic}. This high quality data will assist in the detection of targets along the seafloor. In order to increase the accuracy and precision of the detection process, segmentation of the seafloor texture types will play a key role. Detecting objects of interest within SAS imagery with multiple seafloor texture types is a difficult problem, but if the SAS imagery is properly segmented into the various seafloor texture types, a detector for each region will perform better. Several segmentation algorithms have been applied to SAS imagery \cite{zare2017possibilistic,karoui2009seabed,cobb2011autocorrelation,cobb2011seabed,cobb2013multi} and achieved promising results. One important component of segmentation algorithms is the features from the data. If the features are not optimal, the segmentation algorithm's performance will suffer. Feature selection is the process of choosing the best subset of features that are representative of the data and will lead to the best performance \cite{guyon2003introduction}. In this work, the Xie-Beni(XB) cluster validity measure\cite{xie1991validity} as well as a variation of the XB index were used to select the best number of features.

\section{Motivation}
\subsection{Previous Approaches}
\label{sec:title}
\subsubsection{Segmentation Algorithms}
K-means clustering was based on the approach developed by MacQueen\cite{macqueen1967some}. K-means does a ``crisp" clustering of the data by assigning each data point to only one cluster. In the SAS imagery, there are regions of transition that will need to be accounted for in the algorithm. To capture the regions of transition, fuzzy C-means clustering \cite{bezdek1981objective} can be used so that each data point can be have some membership in multiple clusters. Fuzzy C-means cluster does have issues with noisy images and image artifacts; as a result, a novel fuzzy factor term, G, was developed in the Fuzzy Local Information C-Means (FLICM) \cite{krinidis2010robust} to create a more robust segmentation algorithm. Within the SAS imagery, there may also be outliers that should not be clustered into the various seafloor textures. In order to identify these outliers, typicality can be used. The Possibilistic Fuzzy C-Means (PFCM) algorithm in \cite{pal2005possibilistic} combines the fuzzy and possbilistic clustering approaches to produce membership and typicality values for the data. FLICM and PFCM both have desirable characteristics that will be useful for SAS image segmentation. The Possibilistic Fuzzy Local Information C-Means (PFLICM) \cite{zare2017possibilistic} fuses both approaches and has been proven to segment SAS imagery well. PFLICM is summarized in section 3.1 of this work.
\subsubsection{Feature Selection}
Feature selection is an important area of study within Machine Learning. The advantages of using a subset of features include improved visualization and understanding of the data, efficient use of measurement and storage systems, decrease computational cost in training, and address the curse of dimensionality \cite{guyon2003introduction}.Feature selection methods can be divided into three main categories: wrapper, filter and embedded. The wrapper method selects the best subset of features by generating a model and evaluating each combination of features through some metric. Two wrapper methods are forward and backward feature selection \cite{kohavi1997wrappers}. The wrapper method is a greedy algorithm and can incur high computational cost, but this method should improve performance as features are added. The filter method evaluates the features before using a model. Some filter methods evaluate features based on relevance and other statistical measures \cite{kohavi1997wrappers,peng2005feature}. The filter method is less computationally expensive but may remove features that will help improve the performance of the model or include features that can cause performance to deteriorate \cite{peng2005feature}. Embedded methods combine both approaches and perform feature selection during the training process\cite{guyon2003introduction,chandrashekar2014survey}. Most feature selection processes use error as the evaluation metric but this is not a possible measure of performance for unsupervised applications. Some work has been done in unsupervised feature selection \cite{mitra2002unsupervised,cai2010unsupervised,tang2018robust}. In this unsupervised feature selection framework, the cluster validity measure, XB, will be used to evaluate how well the features cluster the SAS imagery. The XB index and forward feature selection process are detailed in section 3.2 and 3.3 respectively.
\section{METHODS}
\subsection{PFLICM}
\label{sec:title}

The Possibilistic Fuzzy Local Information C-Means (PFLICM) algorithm \cite{zare2017possibilistic} was used to segment the SAS imagery. PFLICM integrates two previous clustering algorithms, the Fuzzy Local Information C-Means (FLICM)\cite{krinidis2010robust} and the Possibilisitic Fuzzy C-Means (PFCM) \cite{pal2005possibilistic}. PFLICM achieves four aims for SAS image segmentation: 1) account for gradients and regions of transition within the images, 2) identify outliers (i.e. targets, clutter), 3) include local spatial information, and 4) minimize computational cost. The objective function for PFLICM is shown in (1): \begin{equation}
J = \sum_{c = 1}^{C}\sum_{n=1}^{N}a{u_{cn}^m\big({||\textbf{x}_n-\textbf{c}_c||}^2_2} + G_{cn}\big) + bt_{cn}{||\textbf{x}_n-\text{c}_c||^2_2} + \sum_{c = 1}^{C}\gamma_c\sum_{n=1}^{N}\big{(1-t_{cn})}^q
\end{equation} subject to the following constraints
\begin{equation}
u_{cn}\geq0 \hspace{10mm}\forall n = 1,...,N \hspace{10mm}\sum_{c=1}^{C}u_{cn} = 1
\end{equation} where $u_{cn}$ is the membership of the $n^{th}$ pixel in the $c^{th}$ cluster, $x_n$ is a dx1 vector for the nth pixel, $c_c$ is a $d \times 1$ vector of the $c^{th}$ cluster, $t_{cn}$ is the typicality value of the $n^{th}$ pixel in the $c^{th}$ cluster, $a$, $b$ and $\gamma_c$ are weights on the membership and typicality terms respectively, and $m$ and $q$ control the ``fuzziness" of the membership values for each cluster and identification of outliers in the data respectively. The $G_{cn}$ term follows from \cite{krinidis2010robust} and incorporates local spatial information:
\begin{equation}
G_{cn} = \underset{k\neq n}{\sum_{{k\in\mathcal{N}_n}}}\frac{1}{d_{nk}+1}\big(1-u_{ck})^m||x_k-c_c||^2_2
\end{equation}
where $\textbf{x}_n$ is the center pixel of a local window, $\mathcal{N}_n$ is the neighborhood around the center pixel, and $d_{nk}$ is the Euclidean distance between the center pixel and one of the neighboring pixels ($x_k$).\\
The cluster centers, membership and typicality values are found via alternating optimization. After these values are randomly initialized, the partial derivative with respect to the variables of interest is taken and set equal to 0 in order to produce three update equations. For the update equation for the membership values, a Lagrange multiplier term was added to enforce the sum-to-one constraint:
\begin{equation}
c_c = \sum_{n}\cfrac{\big(au_{cn}^m + bt_{cn}^q\big)\textbf{x}_n}{\big(au_{cn}^m + bt_{cn}^q\big)}
\end{equation}
\begin{equation}
u_{cn} = \cfrac{1}{\sum_{k=1}^{C}\bigg(\cfrac{(\textbf{x}_n-c_c)(\textbf{x}_n-c_c)^T+G_{cn}}{(\textbf{x}_n-c_k)(\textbf{x}_n-c_k)^T+G_{kn}}\bigg)}
\end{equation}
\begin{equation}
t_{cn} = \cfrac{1}{1+\bigg(\frac{b}{\gamma_c}||x_k-c_c||^2_2\bigg)^{\frac{1}{q-1}}}
\end{equation}
$\gamma_c$ is the mean of the separation of all the data points in the corresponding cluster ($||x_k-c_c||^2_2$). The fuzzy factor term, G, is treated as a constant in the membership update equation despite being a function of neighboring pixels. The actual derivative is not solvable\cite{celik2013comments} but using values from the past iterations produce good segmentation results. In order to reduce the run time of the algorithm, the image is first segmented using superpixels \cite{achanta2012slic} and features are computed from the superpixels. After the segmentation is performed, PFLICM is implemented on the feature vector of each superpixel rather than pixel by pixel.   
\subsection{Cluster Validity Measure}
Clustering is the process of grouping data into various group or clusters that share similar characteristics. Two important metrics for cluster validity is compactness and separation. Compactness refers to how similar the data points are within a cluster while separation measures how distinct the clusters are to one another [14]. In order to look at both metrics, Xie-Beni(XB) cluster validity measure\cite{xie1991validity} used in this work. The XB index is the ratio of the compactness to the separation of the clusters:
\begin{equation}
XB \hspace{.9mm}index = \cfrac{\sum_{i = 1}^{C}\sum_{j = 1}^{N}u_{ij}^2||x_k-c_i||^2_2}{N\cdot\underset{i\neq k}{\min}||c_i-c_k||^2_2}
\end{equation}
The clusters that are well compact and separated should minimize the XB index. The XB index will be used as the evaluation metric to quantify how well segmented the SAS imagery based on each of the extracted features. The original XB index does not take into account the typicality of the data. The extended XB index (EXB) in \cite{zhang2008validity} incorporates typicality by adding the membership and typicality values in the calculation of the index score. For this work, a variation of the XB index (VXB) is proposed
\begin{equation}
VXB\hspace{.9mm} index = \cfrac{\sum_{i = 1}^{C}\sum_{j = 1}^{N}(u_{ij}\cdot t_{ij})^2||x_k-c_i||^2_2}{N\cdot\underset{i\neq k}{\min}||c_i-c_k||^2_2}
\end{equation}
With the VXB, if the data point is an outlier, the membership and typicality values will be low while $||x_k-c_i||^2_2$ will be high. If the data point has high membership with the cluster, then the typicality value will be high while $||x_k-c_i||^2_2$ will be low. The VXB index is used to compare quantitatively PFLICM and PFCM because both algorithms incorporate possiblistic clustering. 
\subsection{Forward Feature Selection}
The selection of the best features for the PFLICM algorithm is crucial to the segmentation process.  A wrapper method, forward feature selection, was implemented in this work. After the features are extracted, one feature is selected. The selected feature is then used to cluster the data with PFLICM and the membership, typicality and cluster center values are estimated. The XB index is then used to score the segmentation produced. The next feature is then selected and the process is repeated until all features have been evaluated individually. The features are then ranked in descending order of the XB index and the best feature is removed from the original feature set and paired with the remaining features. The forward feature selection process will end once all features are exhausted from the original dataset. Figure 1 illustrates the forward feature selection process.
  \begin{figure} [ht]
	\begin{center}
		\begin{tabular}{c} 
			\hspace{-.8mm} 
			\subfloat{
				\includegraphics[scale=.9]{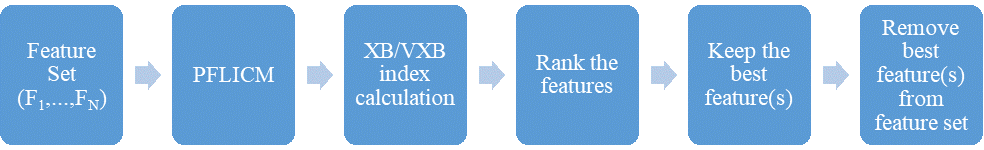}
				\label{fig:fhisthaups}
			}
\hspace{10cm}\\
			\subfloat{
				\includegraphics[scale=.135]{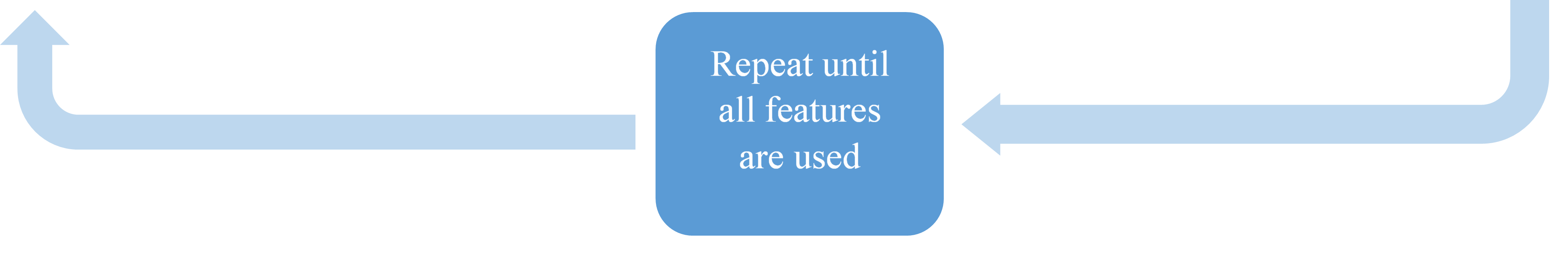}
				\label{fig:fhisthaups}
			}
		\end{tabular}
	\end{center}
	\caption[example] 
	{ \label{fig:example} 
		Forward Feature Selection Process}
\end{figure} 
The suite of features used for the selection process are Sobel \cite{frigui2009detection}, histogram of oriented gradients \cite{dalal2005histograms}, local binary patterns \cite{guo2010completed}, mean, variance, shape\cite{abraham2002novel,cobb2007situ}, Haralick texture features (contrast,correlation,energy, and homogeneity) \cite{haralick1973textural}, a Gabor filter \cite{jain1991unsupervised}, a Gaussian filter \cite{malik2001contour}, lacunarity \cite{williams2015fast}, and a Laplacian of Gaussian filter \cite{malik2001contour,cobb2013multi}. An additional test for the forward feature selection process involved randomly selecting three features from a set of six features (histogram of oriented gradients, homogeneity, lacunarity, local binary patterns, shape, and Sobel) and comparing the randomly selected features with the three features chosen by the XB and VXB indices with PFLICM.
\section{EXPERIMENTAL RESULTS}

The forward feature selection process was implemented on the same image for three sets of features. A grid search was performed to find the optimal parameters that minimized the XB and VXB indices. Table 1 shows the initial value, final value and increments of the parameters used in the grid search. The weight of the typicality was set to be a tenth of the weight on the membership, i.e., $b = .1a$.
 \begin{table}[ht]
	\caption{Grid Search Experiment Values} 
	\label{tab:Paper Margins}
	\begin{center}       
		\begin{tabular}{|c|c|c|c|} 
			\hline
			\rule[-1ex]{0pt}{3.5ex}  Parameters & Initial Value & Final Value & Increment\\
			\hline
			\rule[-1ex]{0pt}{3.5ex}  $a$ & 2 & 12& 2  \\
			\hline
			\rule[-1ex]{0pt}{3.5ex}  $b$ & .2 & 1.2 & .2 \\
			\hline
			\rule[-1ex]{0pt}{3.5ex}  $m$ & 1.2 & 3.0 & .3 \\ 
			\hline
			\rule[-1ex]{0pt}{3.5ex}  $q$ & 1.2 & 2.8 & .2\\
			\hline
		\end{tabular}
	\end{center}
\end{table}
\\The results of the grid search are found in Figure 2. For both the XB and VXB indices, the fuzzifier for the membership ($m$) and typicality ($q$) values were varied as the weights $a$ and $b$ were fixed to 8 and .8 respectively. As seen in Figure 2a-d, the values of $m$ and $q$ that minimize the indices values are 1.8 and 2.8 respectively. Figure 2a-d demonstrate how sensitive the scores are to each fuzzifier. As $n$ becomes smaller, the index increases; additionally, if $m$ is higher than 2, the index also becomes larger. Once the fuzzifier terms were found, the next step was to verify that the weights minimized the XB and VXB metrics. The weights, $a$ and $b$, were varied with $b$ set to be a tenth of $a$ and the scores from this experiment are found in Figure 2e. The set of weights that minimized each score with $m$ and $q$ fixed were $a$ = 14 and $b$ = 1.4. The following parameter settings minimized both the XB and VXB: $a$ = 14, $b$ = 1.4, $m$ = 1.8, $q$ = 2.8, $\epsilon$ = 1e-6 (threshold for change in values) and 3 clusters.

\begin{figure}[ht!]
	\centering
	\subfloat[Log of XB]{
		\includegraphics[scale=0.4]{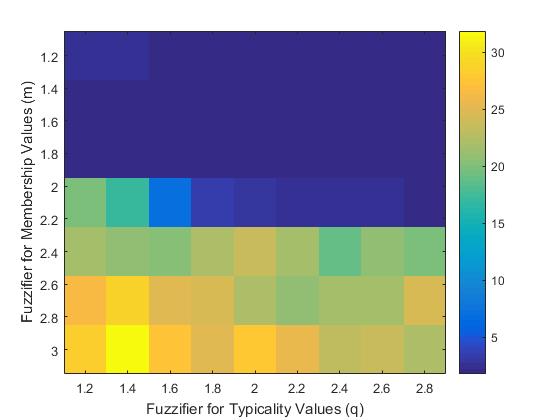}
		\label{fig:fhisthaups}
	}
	~
	\subfloat[Log of VXB]{
		\includegraphics[scale=0.4]{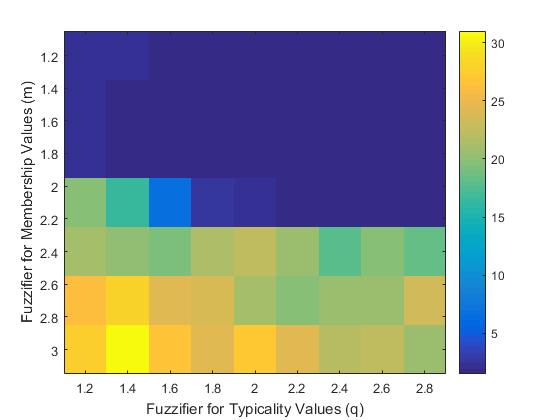}
		\label{fig:histmmps}
	}
	\vskip 0pt
	\subfloat[Log of XB]{
		\includegraphics[scale=0.4]{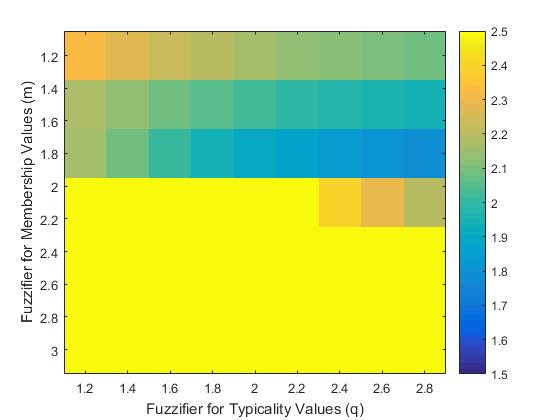}
		\label{fig:fhisthaups2}
	}
	~
	\subfloat[Log of VXB]{
		\includegraphics[scale=0.54]{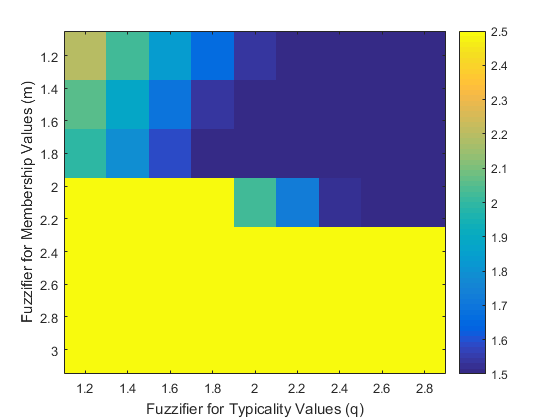}
		\label{fig:histmmps1}
	}
\vskip 0pt
	\subfloat[Log of Indices Scores]{
	\includegraphics[scale=0.54]{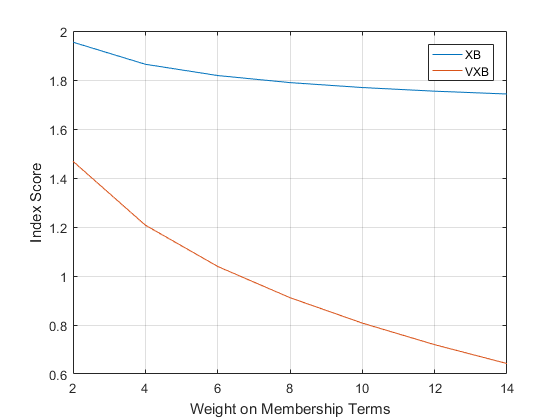}
	\label{fig:histmmps1}
}
	
	\caption{Visualizations of grid search for optimal parameters.  The log of the XB and VXB scores were calculated while m and q were varied. The weights, a and b, were fixed to 8 and .8. (a-b) are scaled between 1.5 and 31 and (c-d) are scaled between 1.5 and 2.5. (e) is the plot of the log of the XB and VXB scores as the weights are varied and m and q are fixed.}
	\label{fig:mildps}
\end{figure}
	
\clearpage

\hspace{-5mm}For each clustering algorithm, the average XB and VXB index score over five runs with random initialization values was recorded and used to determine the selected features. Table 2 summarizes the index scores for each algorithm. The lowest for each index score is highlighted in bold and the second lowest score is underlined. The minimum score indicates the lowest value achieved for the optimal number of features selected by each algorithm. The same parameters for PFLICM were used in PFCM. The VXB was computed for only PFLICM and PFCM since FLICM and K-Means do not produce typicality values. 
\begin{table}[ht]
	\caption{Cluster Validity Measure Results ($\pm$ 1 Standard Deviation)}  
	\label{tab:Paper Margins}
	\begin{center}       
		\begin{tabular}{|c|c|c|c|c|} 
			\hline
			\rule[-1ex]{0pt}{3.5ex}  Index & PFLICM & FLICM & PFCM &K-Means\\
			\hline
			\rule[-1ex]{0pt}{3.5ex}   Min XB & 0.433$\pm$0.158 &\underline{0.405$\pm$0.031}&0.538$\pm$0.070&\textbf{0.284$\pm$3.543e-5} \\
			\hline
			\rule[-1ex]{0pt}{3.5ex}   Min VXB &  \textbf{0.134$\pm$0.035} &-&\underline{0.153$\pm$0.017}&-\\
			\hline
			\rule[-1ex]{0pt}{3.5ex}   XB (all features)  & \underline{1.722$\pm$2.483e-16} &\textbf{1.509$\pm$0.000}&10.994$\pm$0.000&2.089$\pm$1.172e-5\\
			\hline
			\rule[-1ex]{0pt}{3.5ex}   VXB (all features) & \textbf{0.565$\pm$0.000} &-&\underline{2.742$\pm$0.000}&- \\
			\hline
		\end{tabular}
	\end{center}
\end{table}
\\The progression of the XB and VXB scores as feature were added is recorded in Table 3 and 4 respectively. As seen in the tables, the scores should decrease as features are added until a minimum is reached (minimum score for each algorithm is highlighted in bold) and the score should begin to increase as the remaining features are added to the segmentation process.

\begin{table}[ht]
	\caption{ XB Index Score Progression ($\pm$ 1 Standard Deviation)}  
	\label{XB Index Score}
	\begin{center}       
		\begin{tabular}{|c|c|c|c|c|c|} 
			\hline
			\rule[-1ex]{0pt}{3.5ex}   Number of Features & PFLICM & FLICM & PFCM &K-Means\\
			\hline
			\rule[-1ex]{0pt}{3.5ex}   1 & 0.678$\pm$0.161 &0.613$\pm$0.138&\textbf{0.538$\pm$0.070}& \textbf{0.284$\pm$3.543e-5}\\
			\hline
			\rule[-1ex]{0pt}{3.5ex}  2 &0.645$\pm$0.196 &0.617$\pm$0.204&0.551$\pm$0.281&0.526$\pm$0.121\\
			\hline
			\rule[-1ex]{0pt}{3.5ex}    5  & \textbf{0.433$\pm$0.158} &0.702$\pm$0.503&0.562$\pm$0.247&0.815$\pm$0.239\\
			\hline
			\rule[-1ex]{0pt}{3.5ex}   6 & 0.470$\pm$0.086 & \textbf{0.405$\pm$0.031}&0.642$\pm$0.188&0.826$\pm$0.373 \\
			\hline
			\rule[-1ex]{0pt}{3.5ex}   10 & 1.016$\pm$0.169 &0.731$\pm$0.457&1.006$\pm$0.432&1.525$\pm$0.711 \\
			\hline
			\rule[-1ex]{0pt}{3.5ex}   15 & 1.722$\pm$2.483e-16 &1.509$\pm$0.000&10.994$\pm$0.000&2.089$\pm$1.172e-5 \\
			\hline
		\end{tabular}
	\end{center}
\end{table}

\begin{table}[ht]
	\caption{VXB Index Score Progression ($\pm$ 1 Standard Deviation)}  
	\label{tab:Paper Margins}
	\begin{center}       
		\begin{tabular}{|c|c|c|} 
			\hline
			\rule[-1ex]{0pt}{3.5ex}   Number of Features & PFLICM & PFCM\\
			\hline
			\rule[-1ex]{0pt}{3.5ex}    1 & 0.296$\pm$0.053 &0.267$\pm$0.082\\
			\hline
			\rule[-1ex]{0pt}{3.5ex}    3 & 0.164$\pm$0.101 & 0.163$\pm$0.062\\
			\hline
			\rule[-1ex]{0pt}{3.5ex}   6  & \textbf{0.134$\pm$0.035} &0.174$\pm$0.041\\
			\hline
			\rule[-1ex]{0pt}{3.5ex}    9 & 0.327$\pm$0.147 &\textbf{0.153$\pm$0.017} \\
			\hline
			\rule[-1ex]{0pt}{3.5ex}   10& 0.440$\pm$0.162 &0.235$\pm$0.134 \\
			\hline
			\rule[-1ex]{0pt}{3.5ex}   15& 0.565$\pm$0.0000 &2.742$\pm$0.000 \\
			\hline
	\end{tabular}
\end{center}
\end{table}
\hspace{-5mm}The segmentation maps of the features selected using the XB index and all the features are shown in Figures 3 and 4 respectively. Figure 5 shows the progression of the segmentation results as features are added to PFLICM. The three randomly selected features vs the XB and VXB indices selection of three features are found in Figure 6. The selected features by the VXB index are shown for PFLICM and PFCM in Figure 7. The membership maps are shown for FLICM and K-Means and the product maps (membership*typicality) for PFLICM and PFCM.

\begin{figure}[ht!]
	\centering
	\subfloat[Original Image]{
		\includegraphics[scale=1.60]{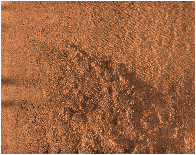}
		\label{fig:fhisthaups}
	}
\vskip 0pt
	\subfloat[PFLICM Cluster 1]{
		\includegraphics[scale=0.3]{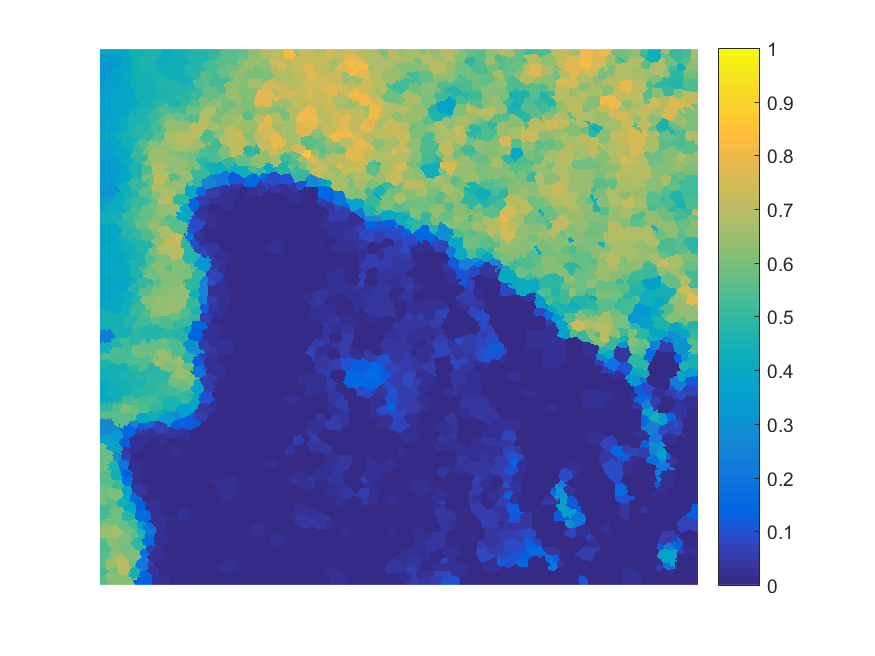}
		\label{fig:fhisthaups}
	}
	~
	\subfloat[PFLICM Cluster 2]{
	\includegraphics[scale=0.3]{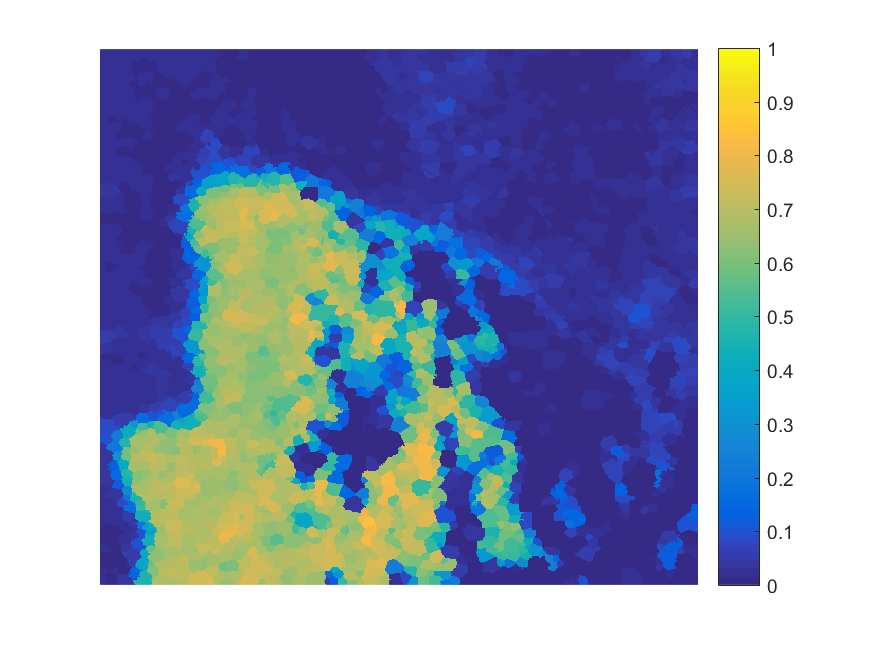}
		\label{fig:histmmps}
	}
	~
	\subfloat[PFLICM Cluster 3]{
	\includegraphics[scale=0.3]{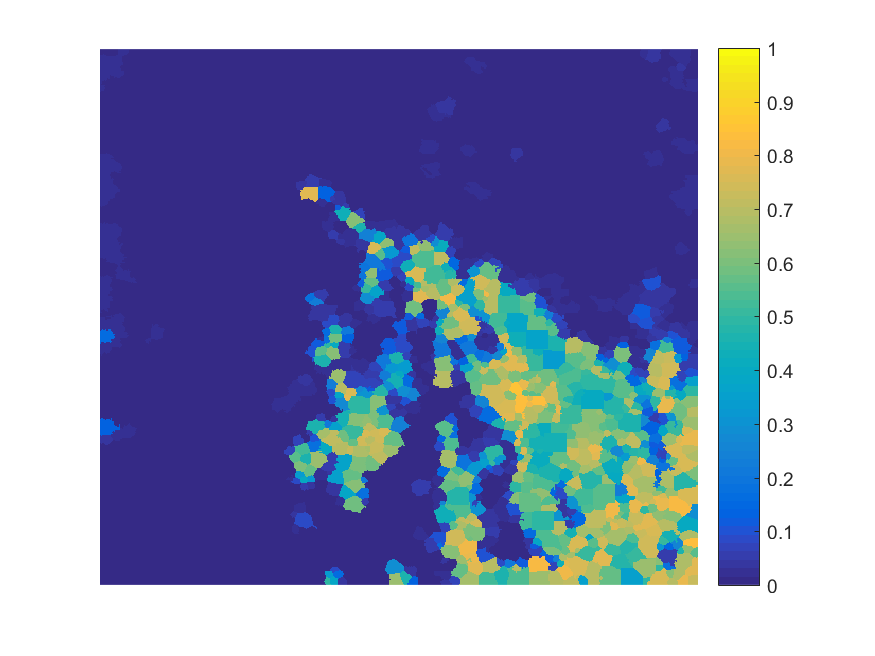}
		\label{fig:histminps}
	}
\vskip 0pt
	\subfloat[PFCM Cluster 1]{
	\includegraphics[scale=0.3]{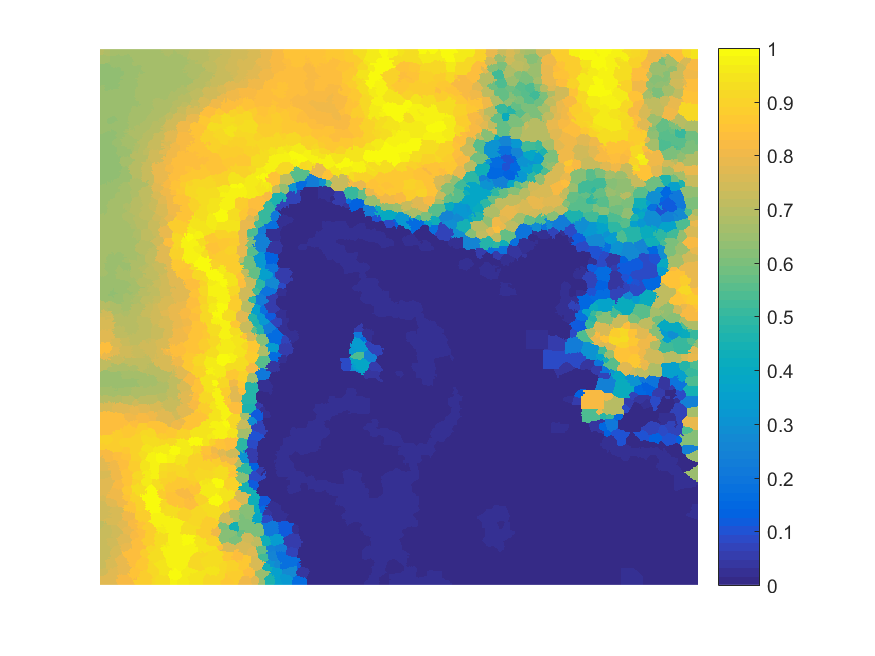}
	\label{fig:fhisthaups2}
}
~
\subfloat[PFCM Cluster 2]{
	\includegraphics[scale=0.3]{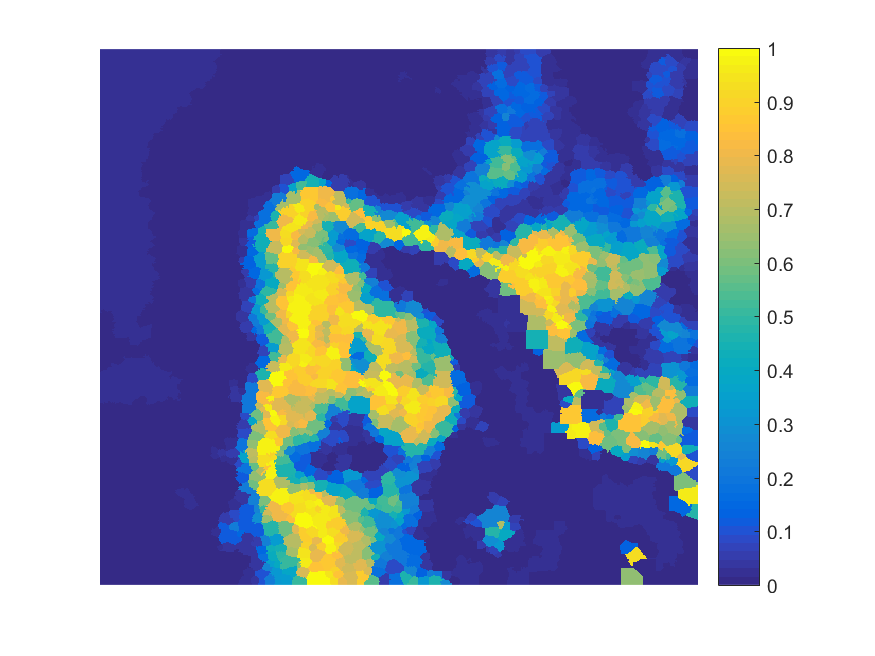}
	\label{fig:histmmps1}
}
~
\subfloat[PFCM Cluster 3]{
	\includegraphics[scale=0.3]{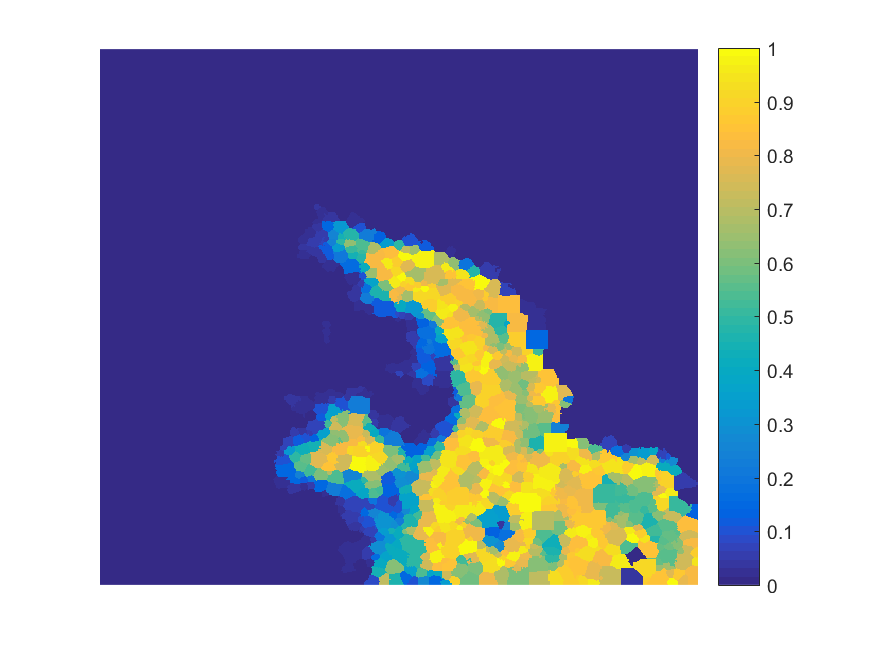}
	\label{fig:histminps2}
}

\subfloat[FLICM Cluster 1]{
	\includegraphics[scale=0.3]{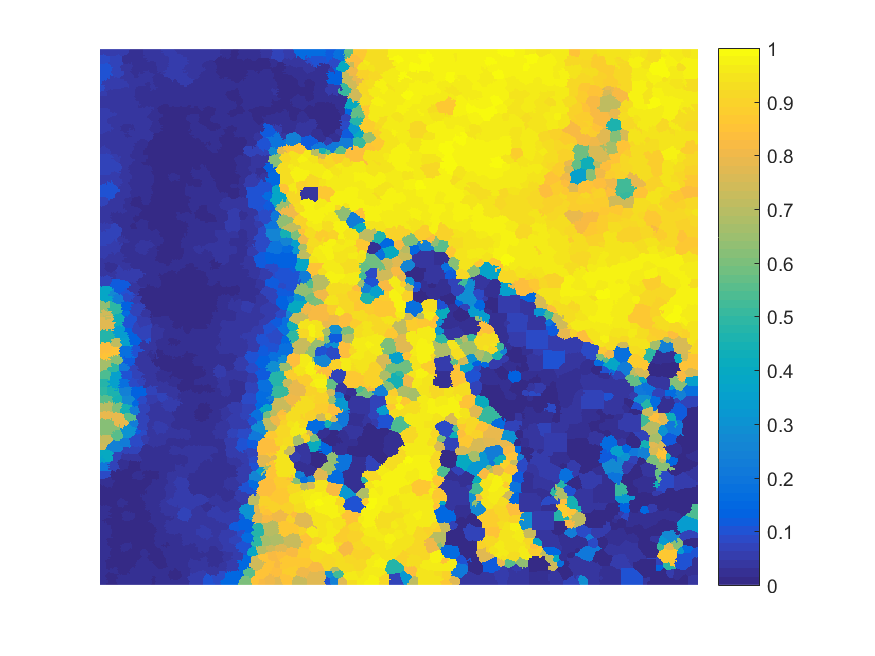}
	\label{fig:fhisthaups4}
}
~
\subfloat[FLICM Cluster 2]{
	\includegraphics[scale=0.3]{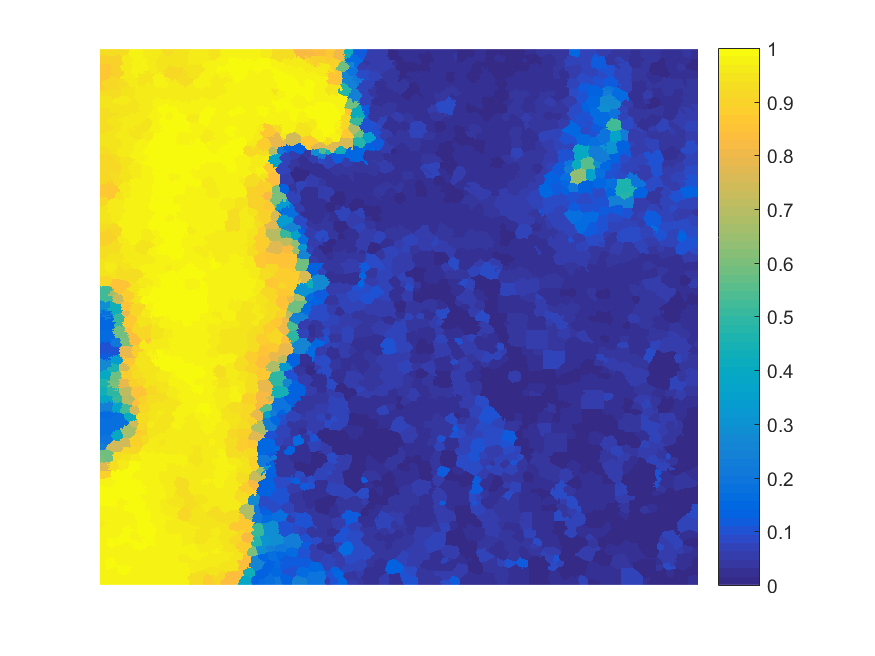}
	\label{fig:histmmps2}
}
~
\subfloat[FLICM Cluster 3]{
	\includegraphics[scale=0.3]{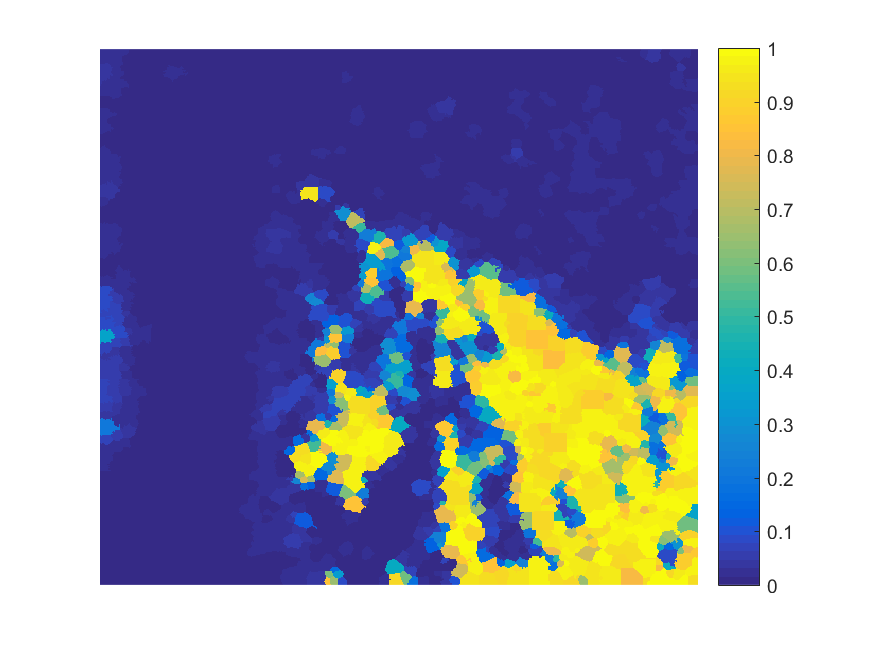}
	\label{fig:histminps3}
}
\vskip 0pt
\subfloat[K-Means Cluster 1]{
	\includegraphics[scale=0.3]{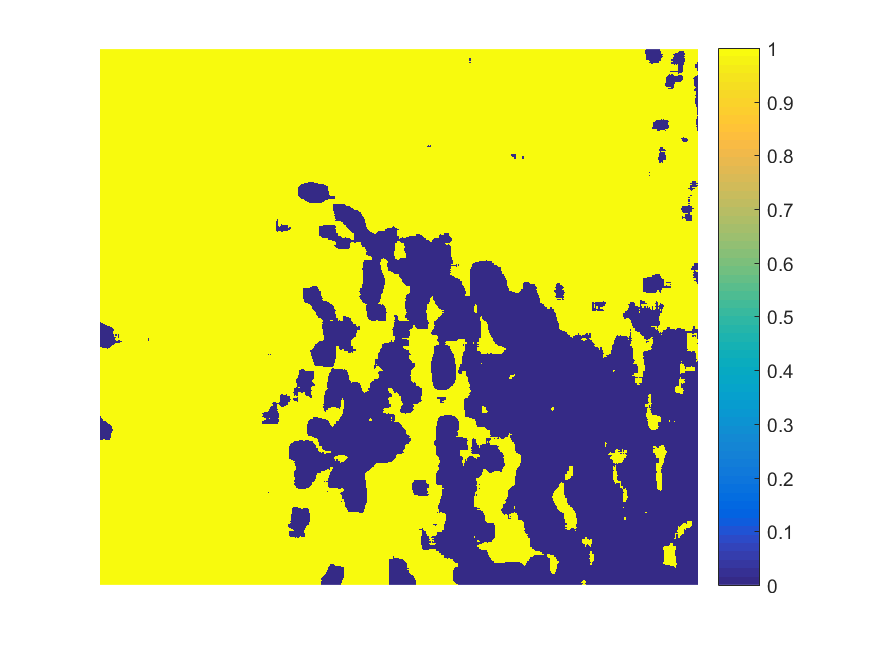}
	\label{fig:fhisthaups4}
}
~
\subfloat[K-Means Cluster 2]{
	\includegraphics[scale=0.3]{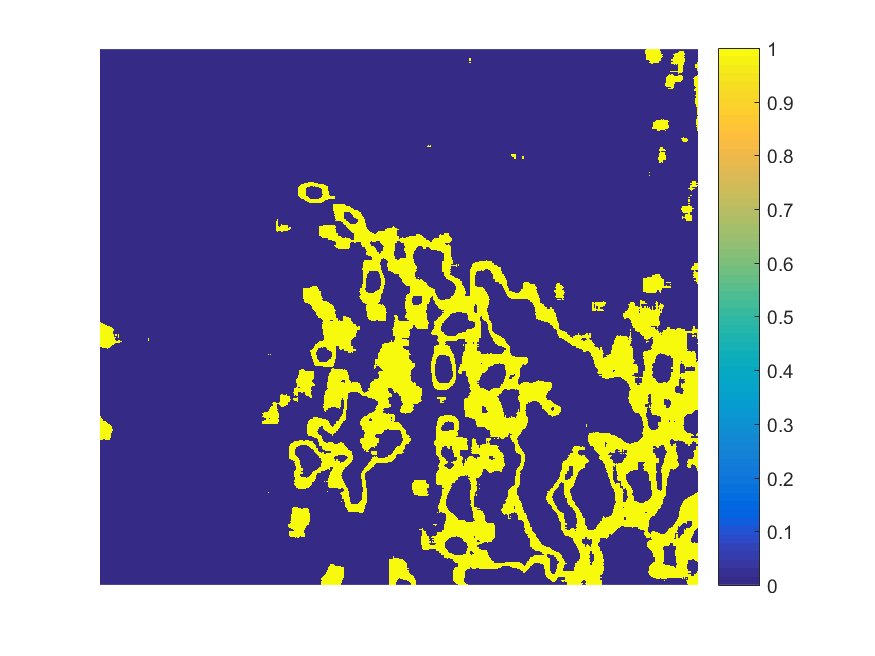}
	\label{fig:histmmps2}
}
~
\subfloat[K-Means Cluster 3]{
	\includegraphics[scale=0.3]{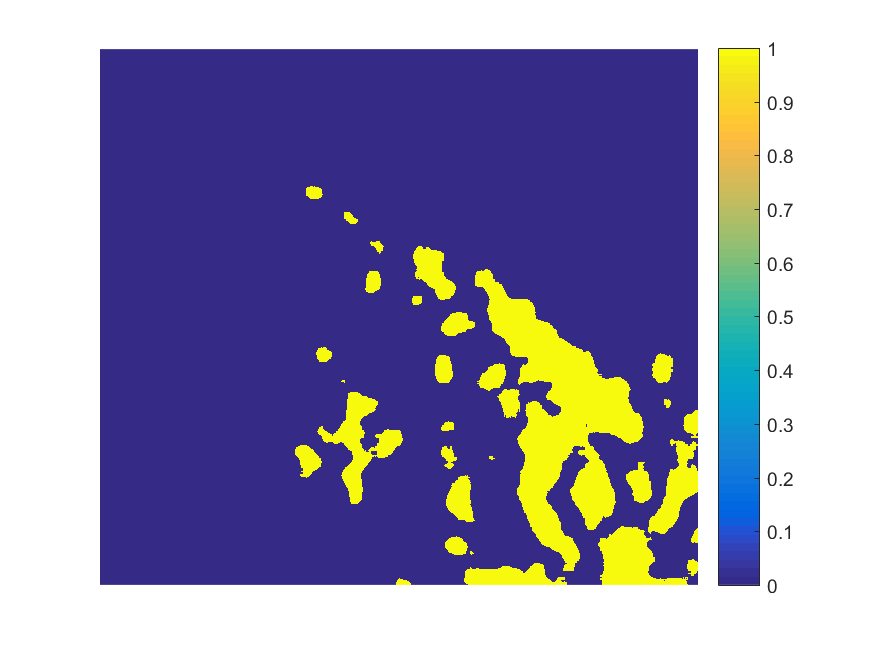}
	\label{fig:histminps3}
}
	\caption{(a) SAS image with distinct regions: sand ripple, hard packed sand, and coral/rocky surface. The segmentation results for PFLICM (b-d), PFCM (e-g), FLICM (h-j), and K-Means(k-m) with the selected features using XB index.}
	\label{fig:mildps}
\end{figure}
\vspace{-10mm}

\begin{figure}[ht!]
	\centering
	\subfloat[Original Image]{
		\includegraphics[scale=1.60]{SPIE_img.png}
		\label{fig:fhisthaups3}
	}
\vskip 0pt
	\subfloat[PFLICM Cluster 1]{
	\includegraphics[scale=0.3]{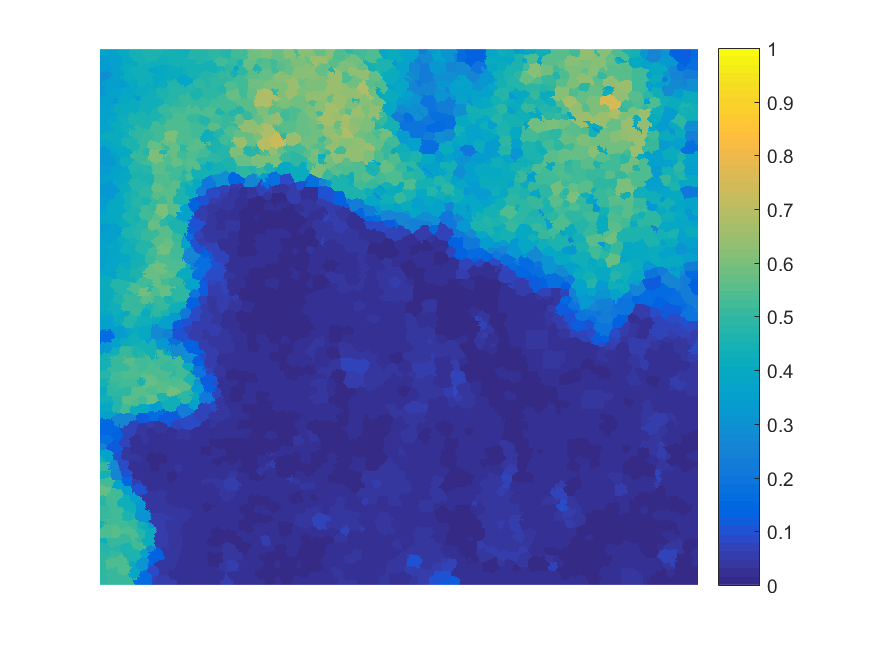}
	\label{fig:fhistha}
}
~
\subfloat[PFLICM Cluster 2]{
	\includegraphics[scale=0.3]{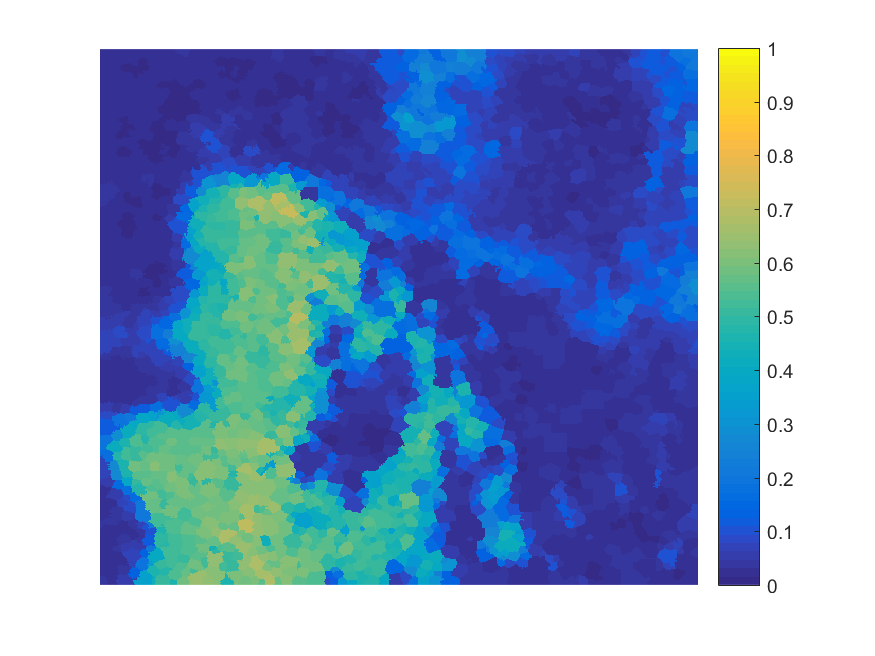}
	\label{fig:histmms}
}
~
\subfloat[PFLICM Cluster 3]{
	\includegraphics[scale=0.3]{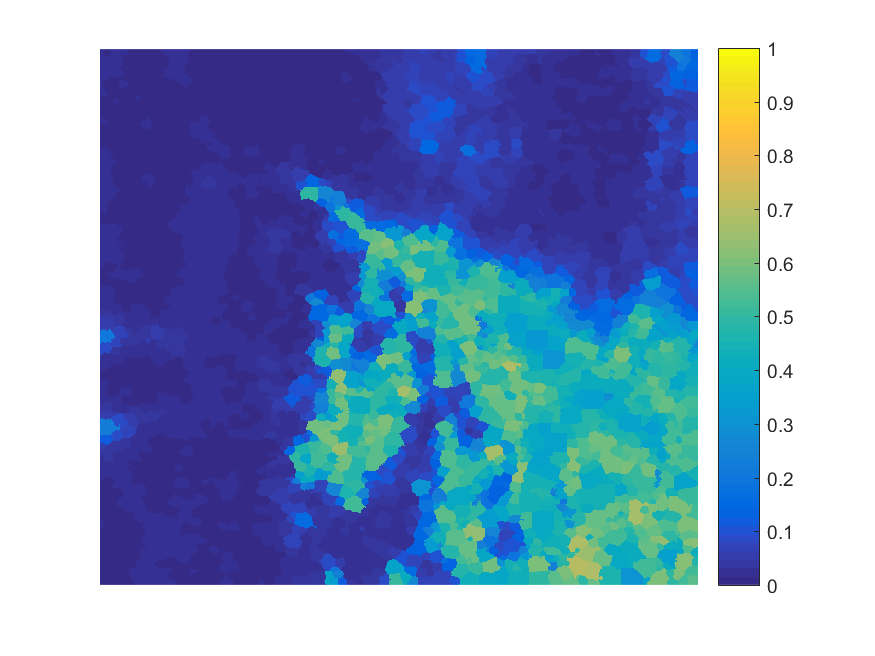}
	\label{fig:histminps}
}
\vskip 0pt
\subfloat[PFCM Cluster 1]{
	\includegraphics[scale=0.3]{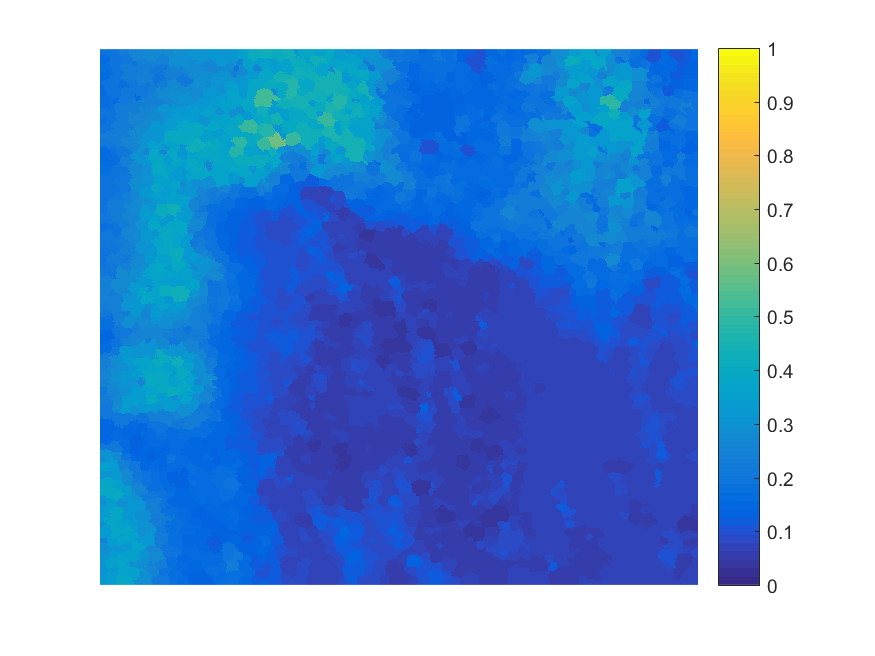}
	\label{fig:fhisthaups2}
}
~
\subfloat[PFCM Cluster 2]{
	\includegraphics[scale=0.3]{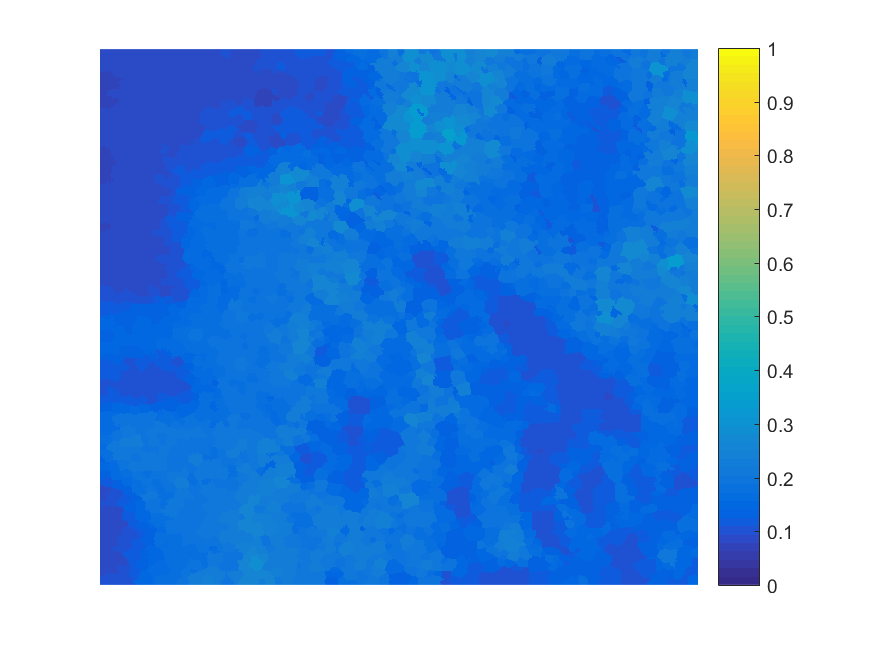}
	\label{fig:histmmps1}
}
~
\subfloat[PFCM Cluster 3]{
	\includegraphics[scale=0.3]{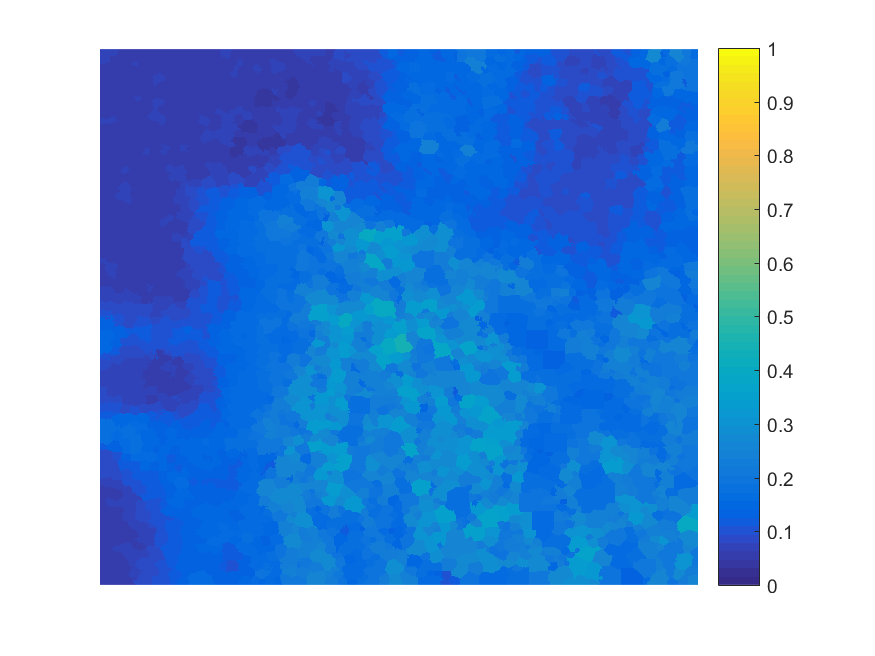}
	\label{fig:histminps2}
}
\vskip 0pt
	\subfloat[FLICM Cluster 1]{
		\includegraphics[scale=0.3]{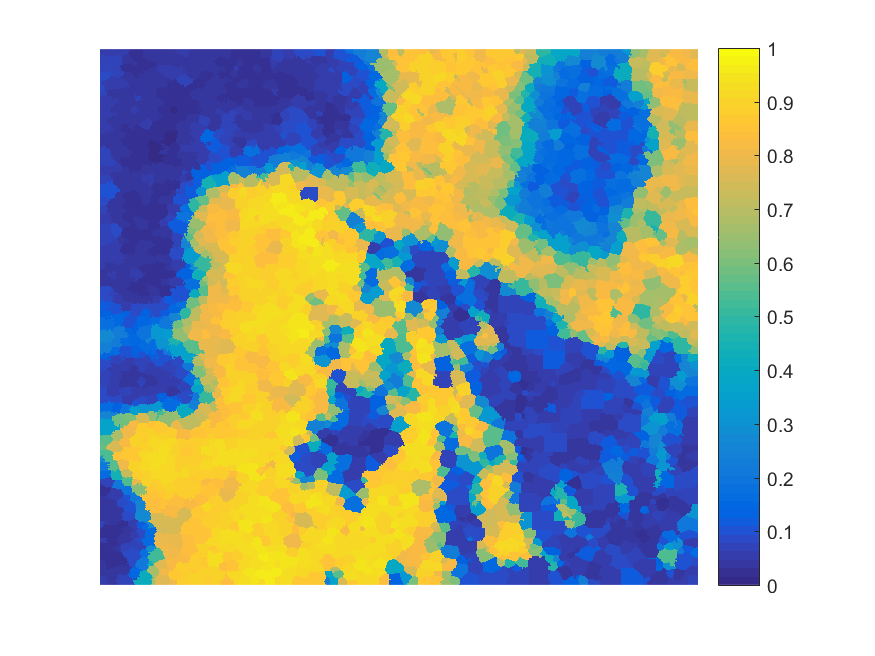}
		\label{fig:histrelps6}
		~
	}
	\subfloat[FLICM Cluster 2]{
		\includegraphics[scale=0.3]{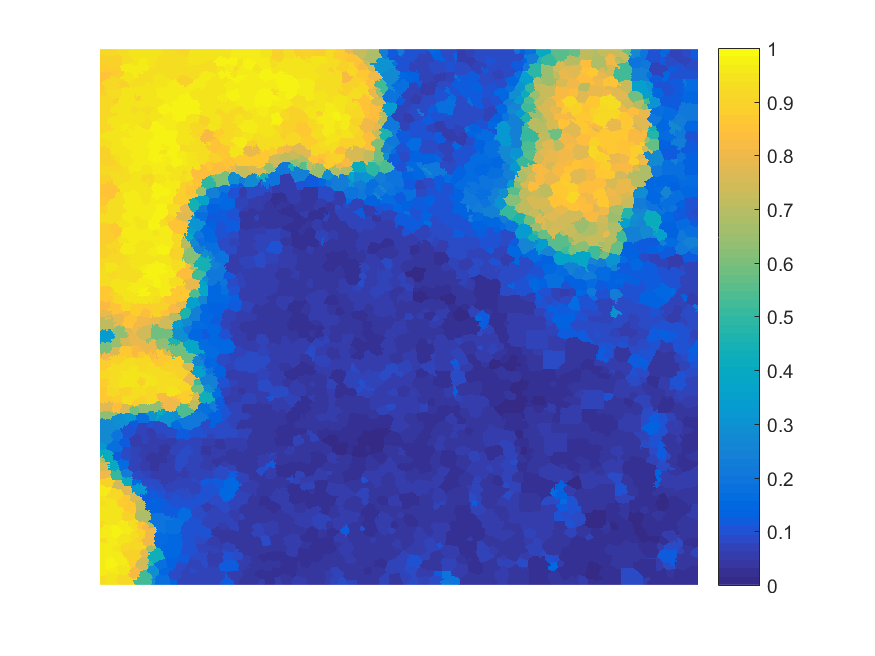}
		\label{fig:histrelps7}
	}
	~
	\subfloat[FLICM Cluster 3]{
		\includegraphics[scale=0.3]{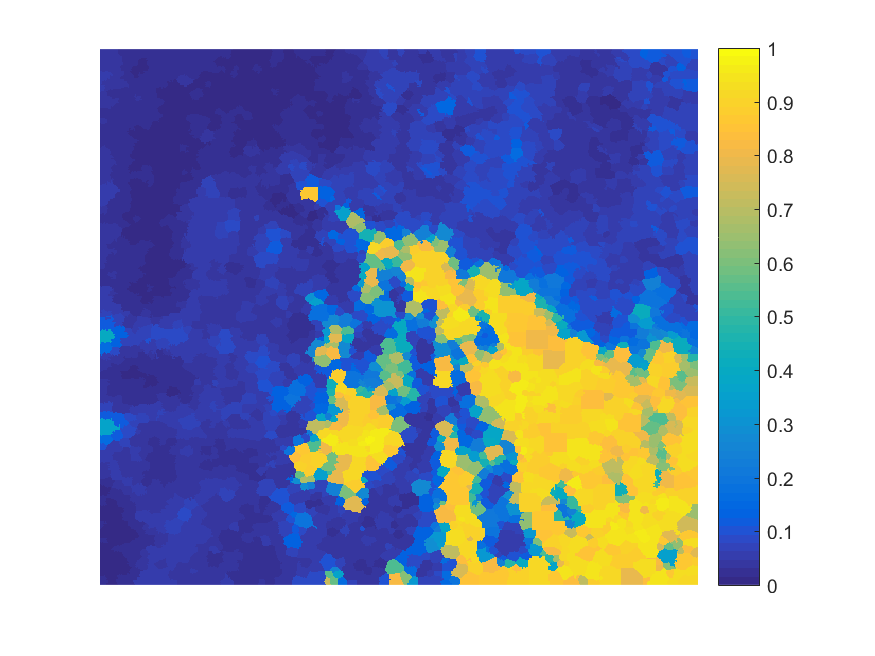}
		\label{fig:histrelps8}
	}
\vskip 0pt
\subfloat[K-Means Cluster 1]{
	\includegraphics[scale=0.3]{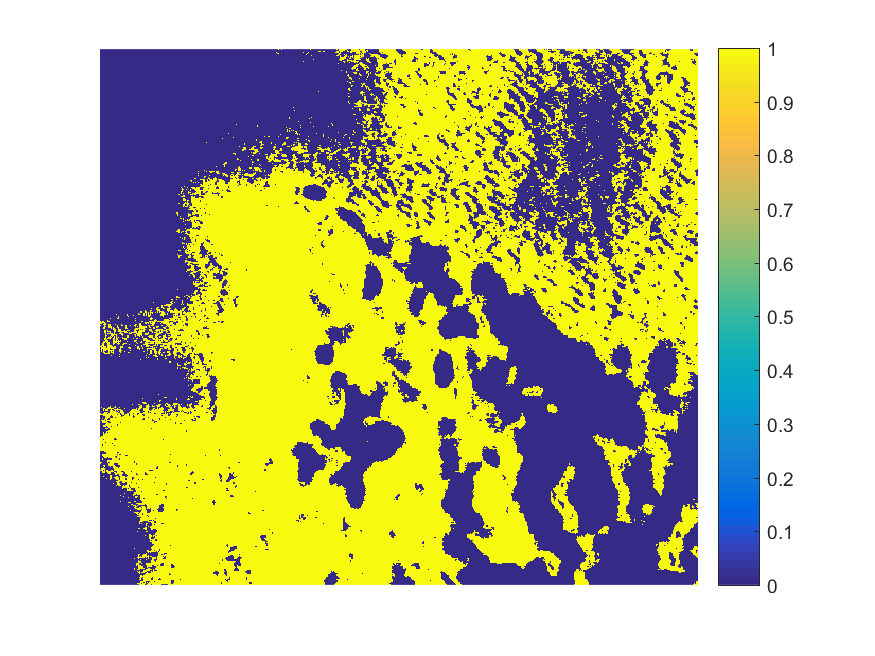}
	\label{fig:fhisthaups4}
}
~
\subfloat[K-Means Cluster 2]{
	\includegraphics[scale=0.3]{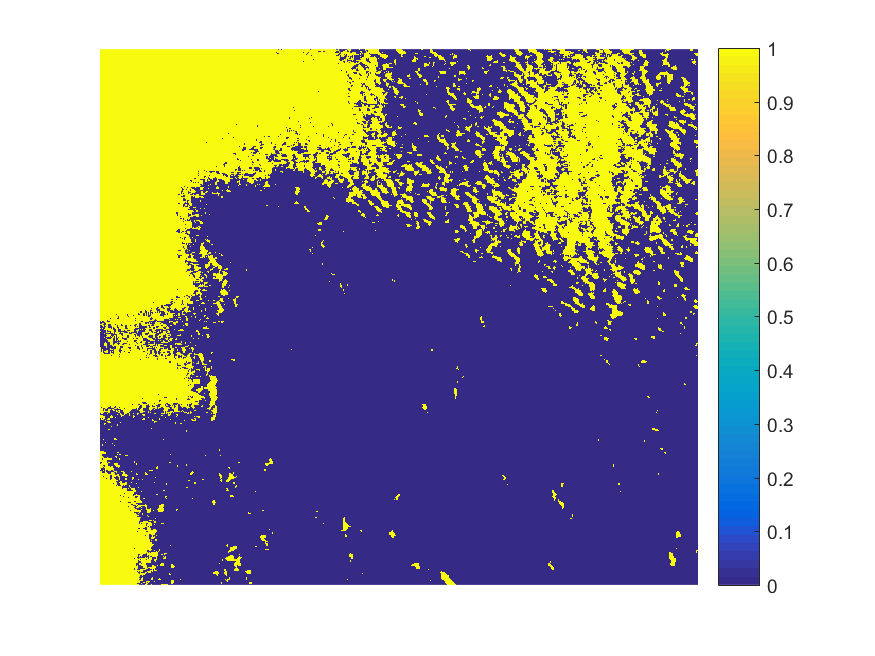}
	\label{fig:histmmps2}
}
~
\subfloat[K-Means Cluster 3]{
	\includegraphics[scale=0.3]{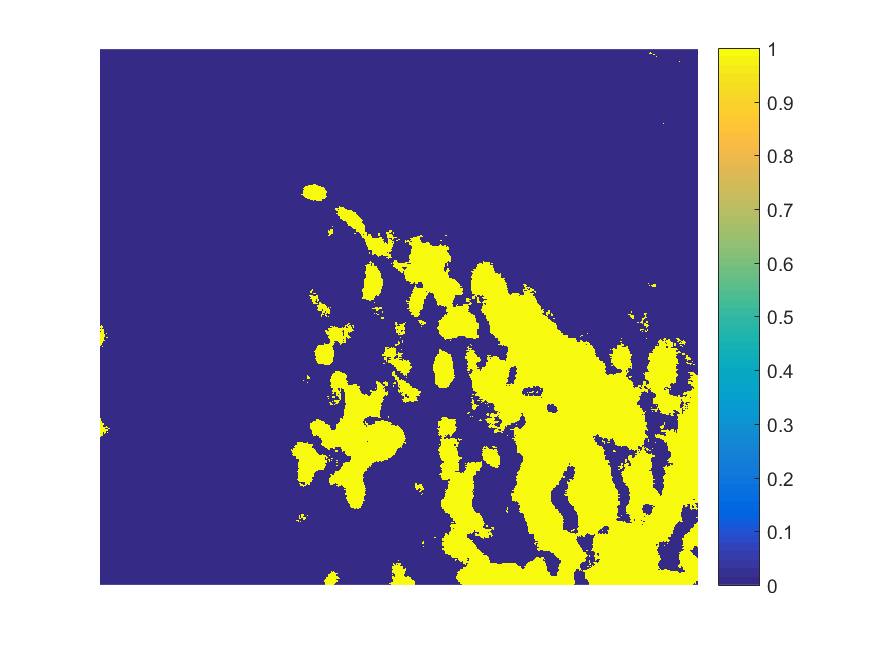}
	\label{fig:histminps3}
}

	\caption{(a) SAS image with distinct regions: sand ripple, hard packed sand, and coral/rocky surface. The segmentation results for PFLICM (b-d), PFCM (e-g), FLICM (h-j), and K-Means(k-m) with all features.}
	\label{fig:mildps1}
\end{figure}
\vspace{-25mm}	
\clearpage
\begin{figure}[ht!]
	\centering
	\subfloat[Original Image]{
		\graphicspath{ {C:/Users/jpeeples/SPIE/} }
		\includegraphics[scale=1.60]{SPIE_img.png}
		\label{fig:fhisthaups}
	}
	\vskip 0pt
	\subfloat[PFLICM Cluster 1]{
		\includegraphics[scale=0.3]{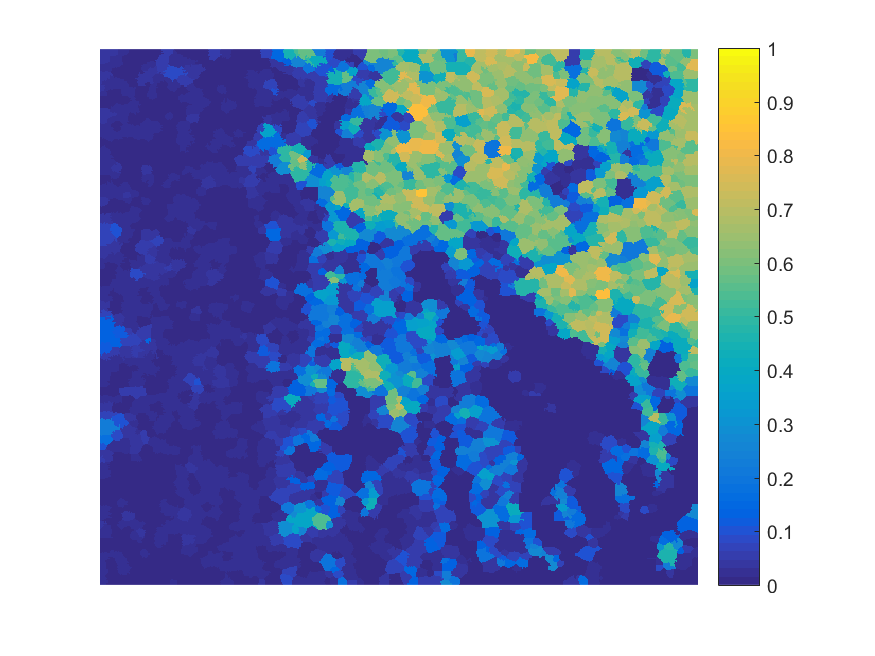}
		\label{fig:fhisthaups}
	}
	~
	\subfloat[PFLICM Cluster 2]{
		\includegraphics[scale=0.3]{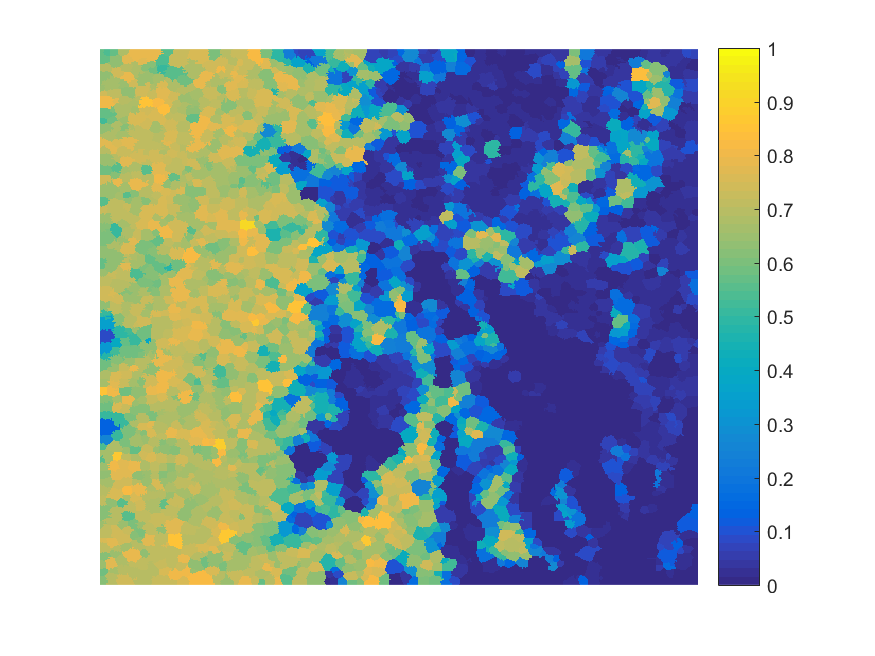}
		\label{fig:histmmps}
	}
	~
	\subfloat[PFLICM Cluster 3]{
		\includegraphics[scale=0.3]{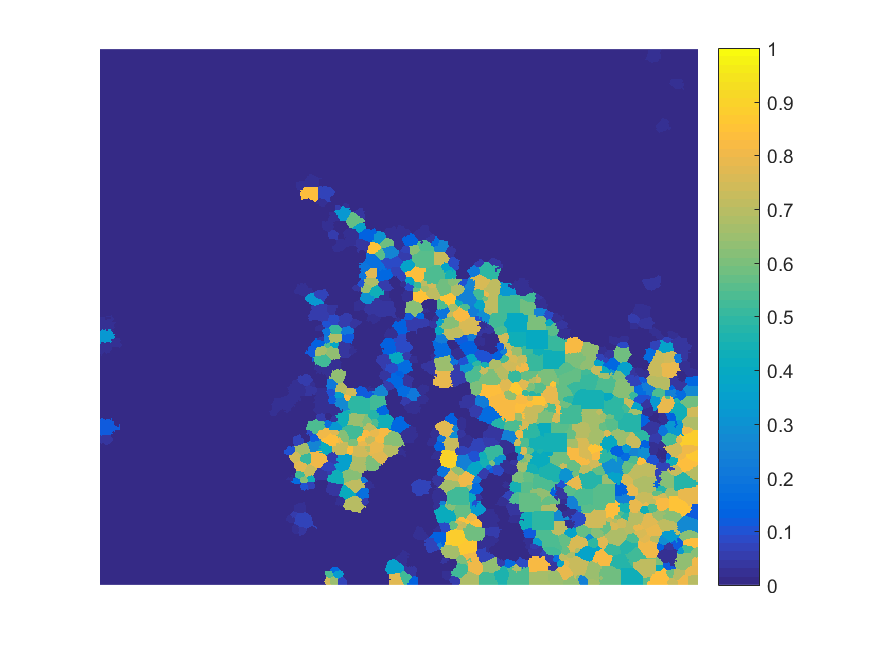}
		\label{fig:histminps}
	}
	
	\vskip 0pt
	\subfloat[PFLICM Cluster 1]{
		\includegraphics[scale=0.3]{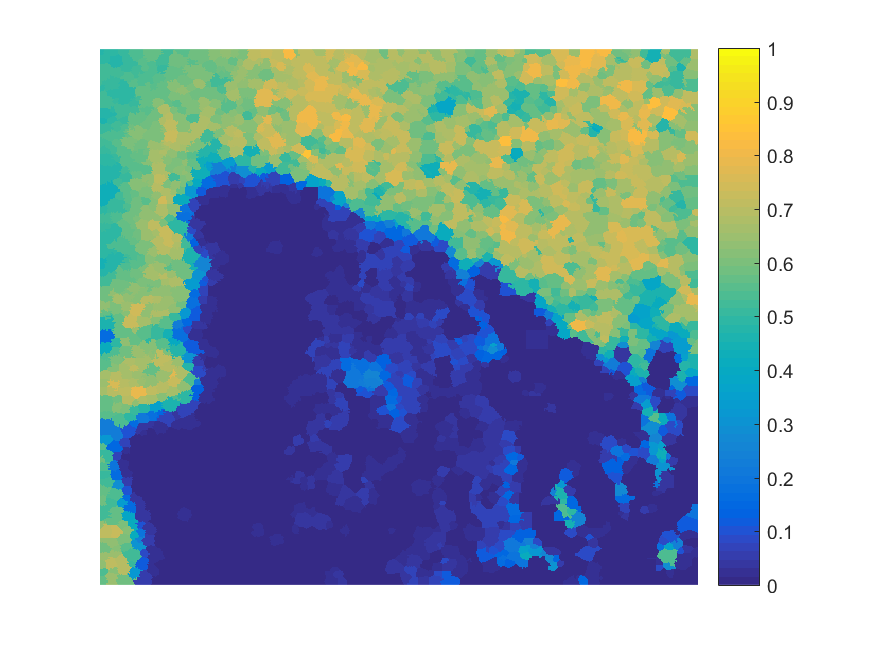}
		\label{fig:fhisthaups2}
	}
	~
	\subfloat[PFLICM Cluster 2]{
		\includegraphics[scale=0.3]{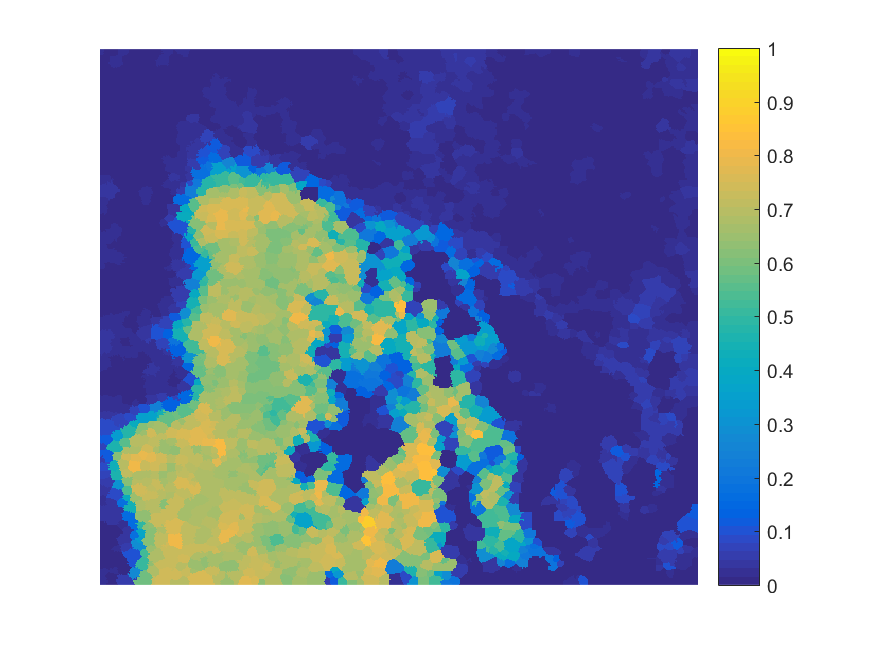}
		\label{fig:histmmps1}
	}
	~
	\subfloat[PFLICM Cluster 3]{
		\includegraphics[scale=0.3]{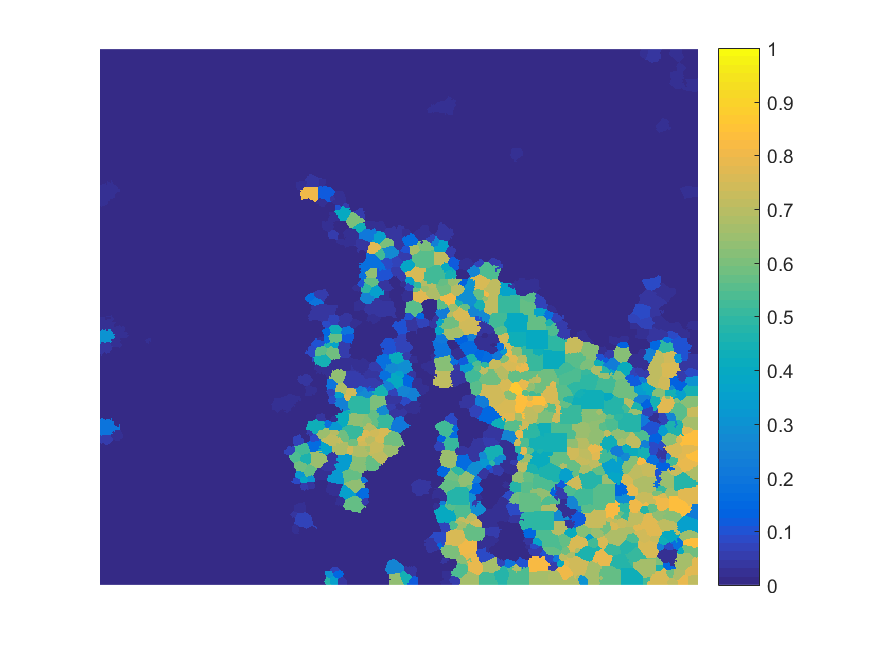}
		\label{fig:histminps2}
	}
	
		\subfloat[PFLICM Cluster 1]{
		\includegraphics[scale=0.3]{1_PFLICM_Product_C2.png}
		\label{fig:fhisthaups}
	}
	~
	\subfloat[PFLICM Cluster 2]{
		\includegraphics[scale=0.3]{1_PFLICM_Product_C1.png}
		\label{fig:histmmps}
	}
	~
	\subfloat[PFLICM Cluster 3]{
		\includegraphics[scale=0.3]{1_PFLICM_Product_C3.png}
		\label{fig:histminps}
	}
	
	\vskip 0pt
	\subfloat[PFLICM Cluster 1]{
		\includegraphics[scale=0.3]{5_PFLICM_Product_C3.png}
		\label{fig:fhisthaups}
	}
	~
	\subfloat[PFLICM Cluster 2]{
		\includegraphics[scale=0.3]{5_PFLICM_Product_C1.png}
		\label{fig:histmmps}
	}
	~
	\subfloat[PFLICM Cluster 3]{
		\includegraphics[scale=0.3]{5_PFLICM_Product_C2.png}
		\label{fig:histminps}
}
	\caption{(a) SAS image with distinct regions: sand ripple, hard packed sand, and coral/rocky surface. The segmentation results for PFLICM with the first feature (b-d), three features (e-g), selected features (h-j), and all features (k-m).}
	\label{fig:mildps}
\end{figure}
\vspace{-25mm}	
\begin{figure}[ht!]
	\centering
	\subfloat[Original Image]{
		\includegraphics[scale=1.6]{SPIE_img.png}
		\label{fig:fhisthaups}
	}
	\vskip 0pt
	\subfloat[PFLICM Cluster 1]{
		\includegraphics[scale=0.3]{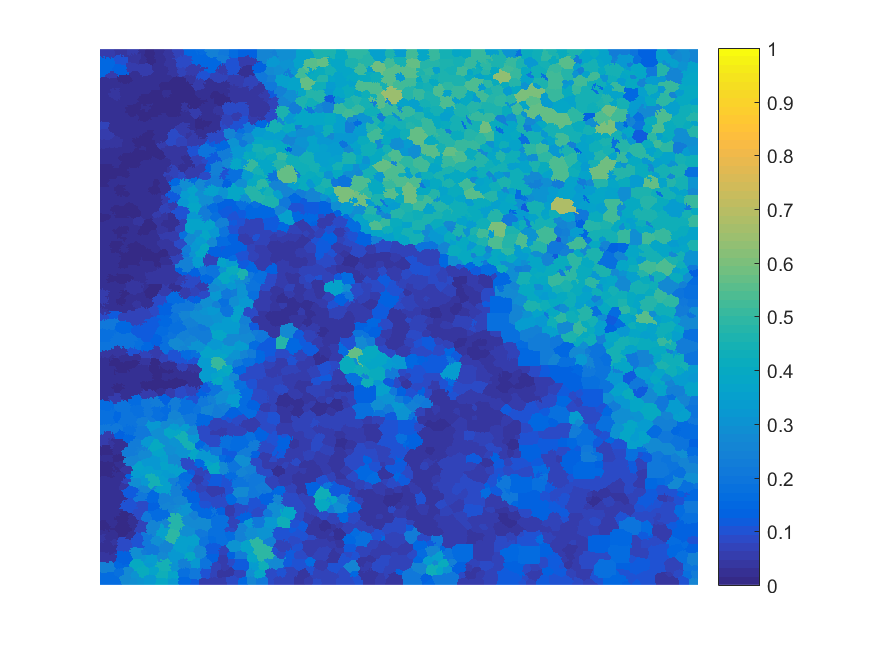}
		\label{fig:fhisthaups}
	}
	~
	\subfloat[PFLICM Cluster 2]{
		\includegraphics[scale=0.3]{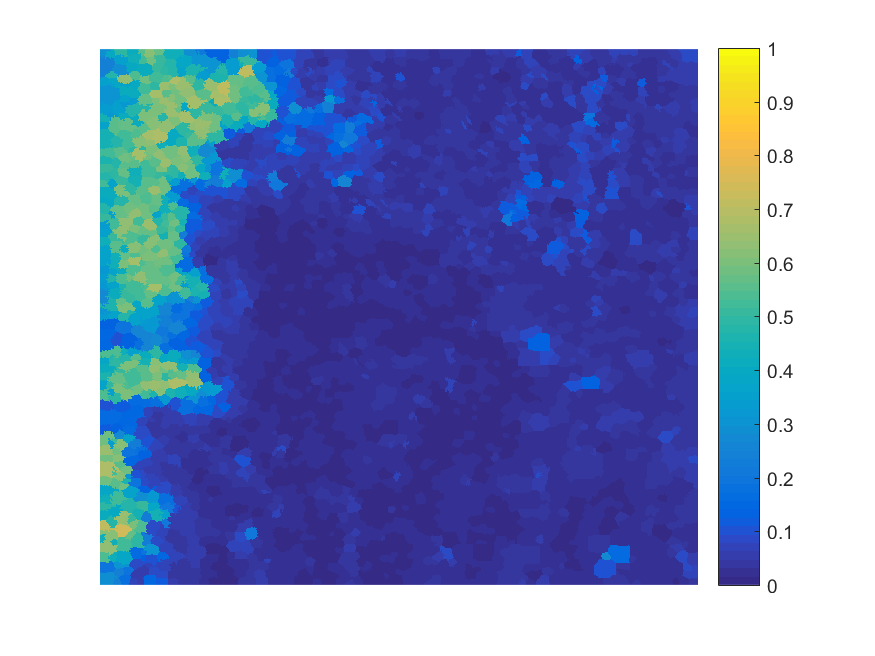}
		\label{fig:histmmps}
	}
	~
	\subfloat[PFLICM Cluster 3]{
		\includegraphics[scale=0.3]{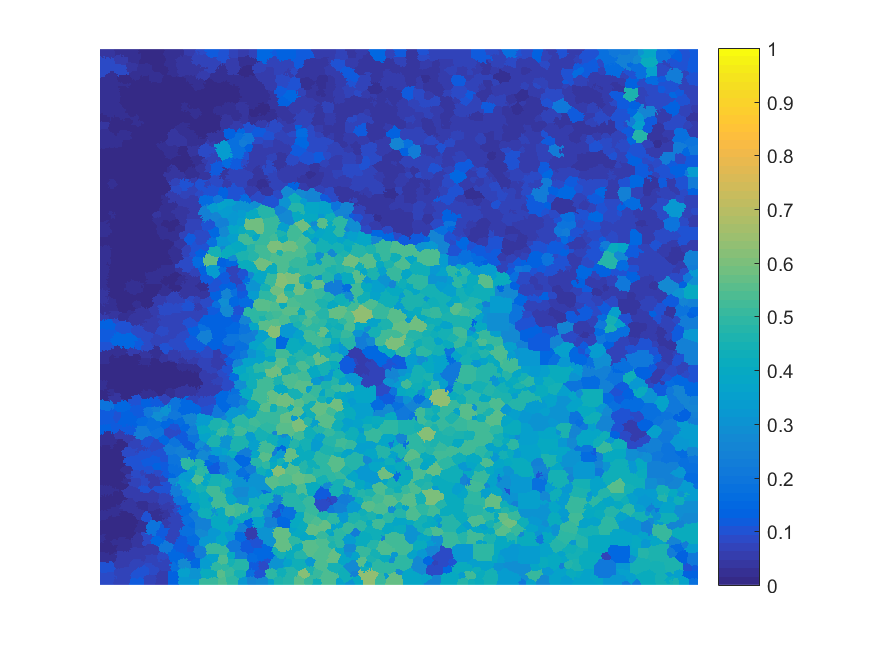}
		\label{fig:histminps}
	}
	
	\vskip 0pt
	\subfloat[PFLICM Cluster 1]{
		\includegraphics[scale=0.3]{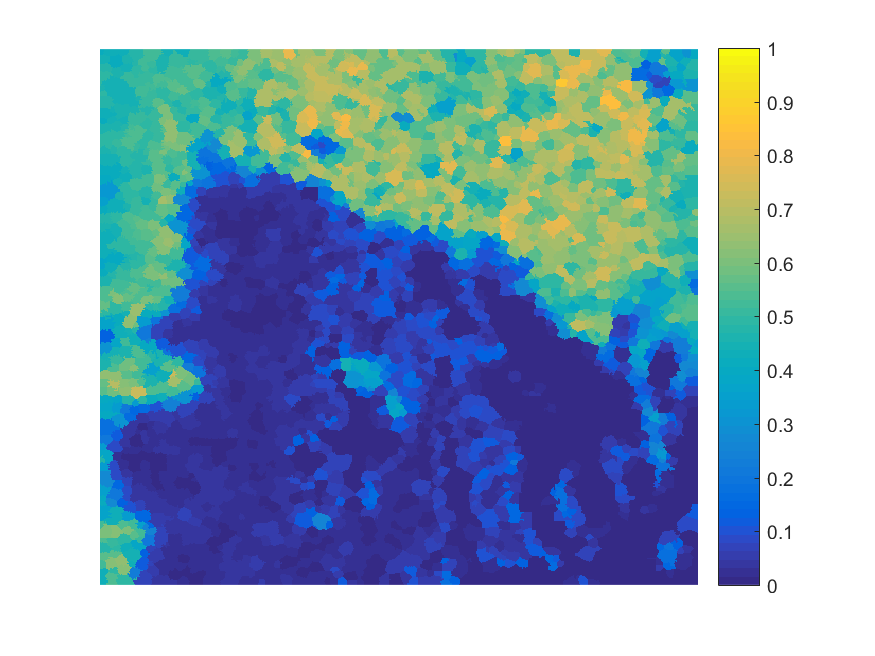}
		\label{fig:fhisthaups2}
	}
	~
	\subfloat[PFLICM Cluster 2]{
		\includegraphics[scale=0.3]{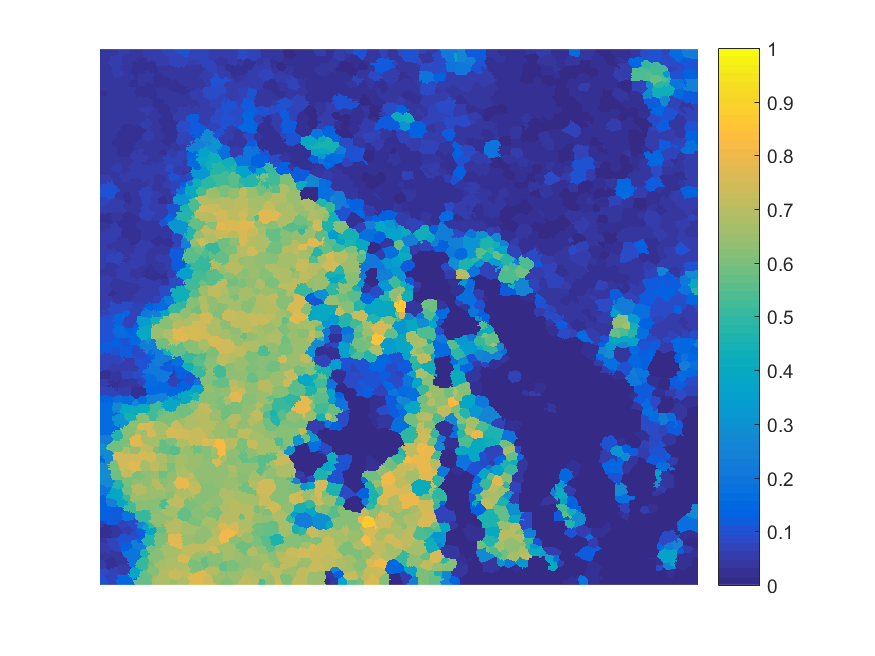}
		\label{fig:histmmps1}
	}
	~
	\subfloat[PFLICM Cluster 3]{
		\includegraphics[scale=0.3]{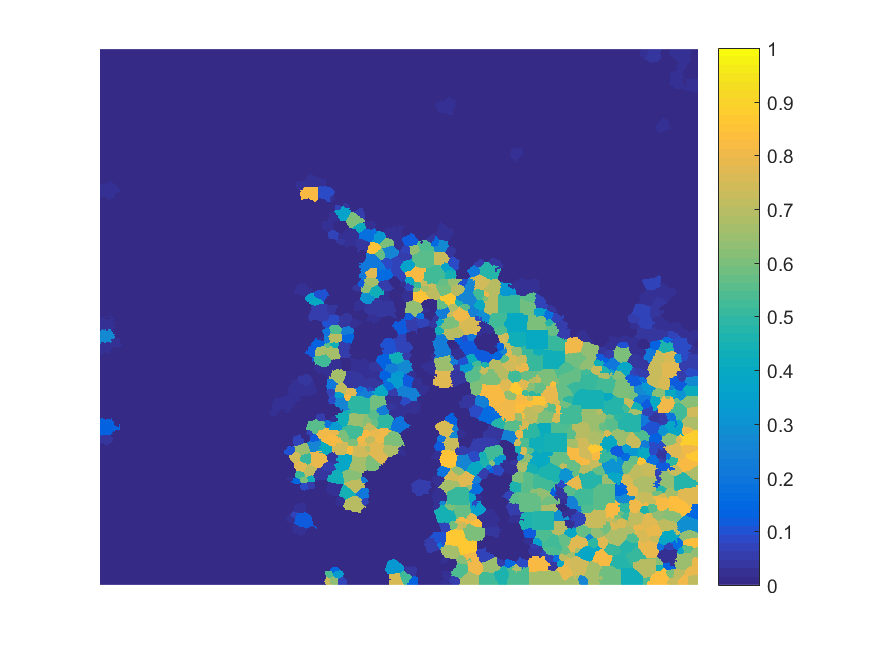}
		\label{fig:histminps2}
	}
\vskip 0pt
\subfloat[PFLICM Cluster 1]{
	\includegraphics[scale=0.3]{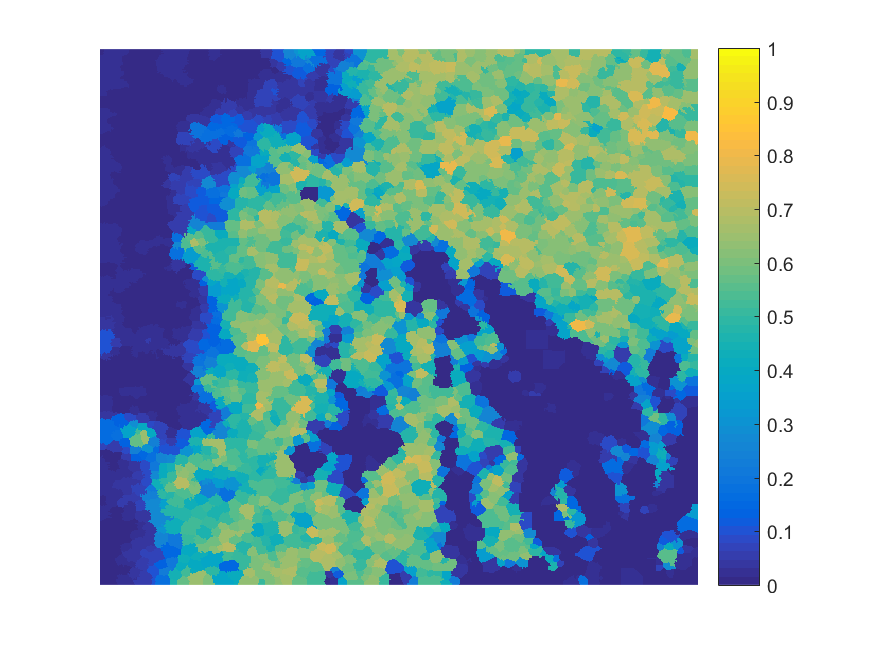}
	\label{fig:fhisthaups}
}
~
\subfloat[PFLICM Cluster 2]{
	\includegraphics[scale=0.3]{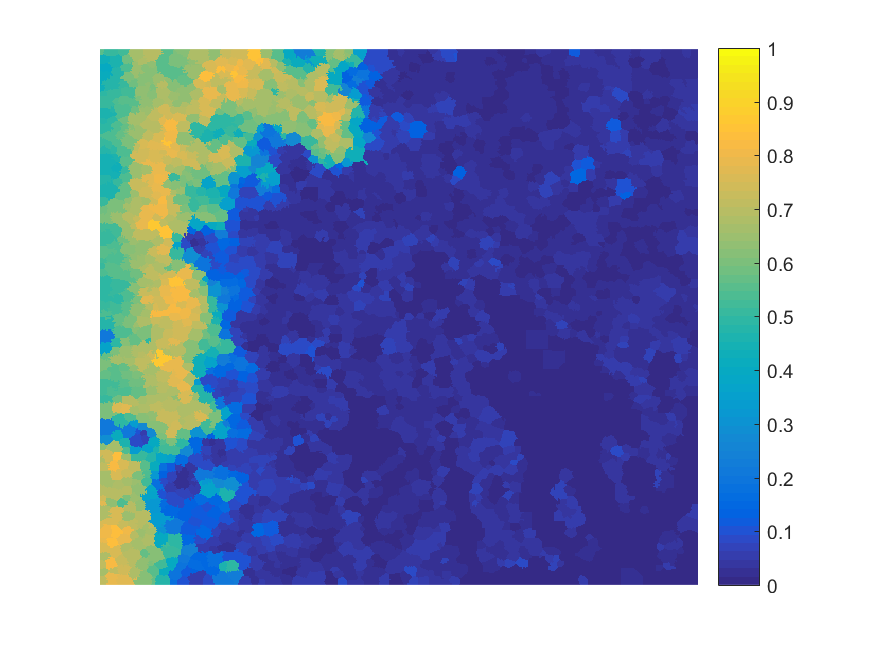}
	\label{fig:histmmps}
}
~
\subfloat[PFLICM Cluster 3]{
	\includegraphics[scale=0.3]{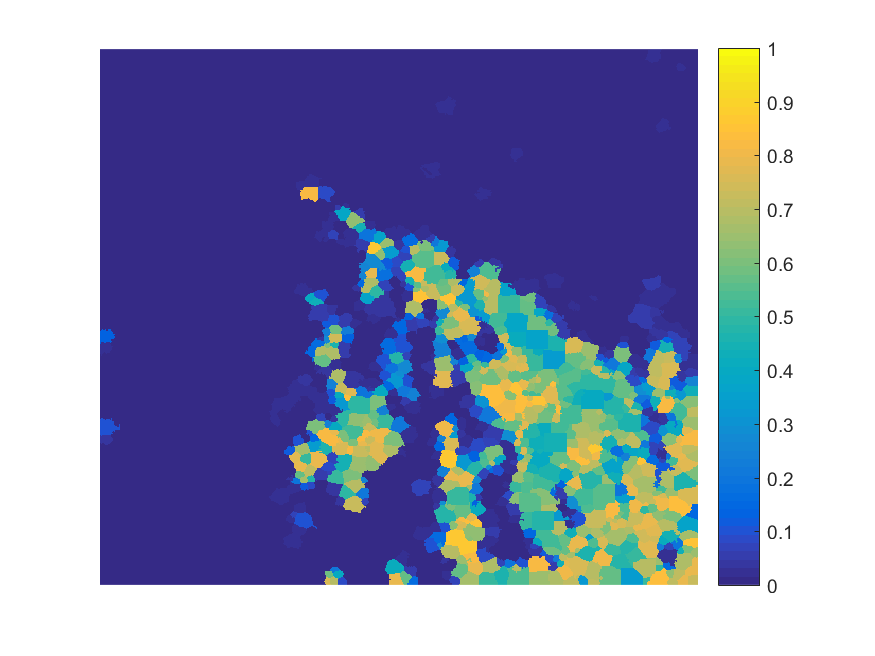}
	\label{fig:histminps}
}

	\caption{(a) SAS image with distinct regions: sand ripple, hard packed sand, and coral/rocky surface. The segmentation results for randomly selected features (b-d), XB selected features (e-g) and VXB selected features (h-j).}
	\label{fig:mildps}
\end{figure}
\clearpage
\begin{figure}[ht!]
	\centering
	\subfloat[Original Image]{
		\includegraphics[scale=1.6]{SPIE_img.png}
		\label{fig:fhisthaups}
	}
	\vskip 0pt
	\subfloat[PFLICM Cluster 1]{
		\includegraphics[scale=0.3]{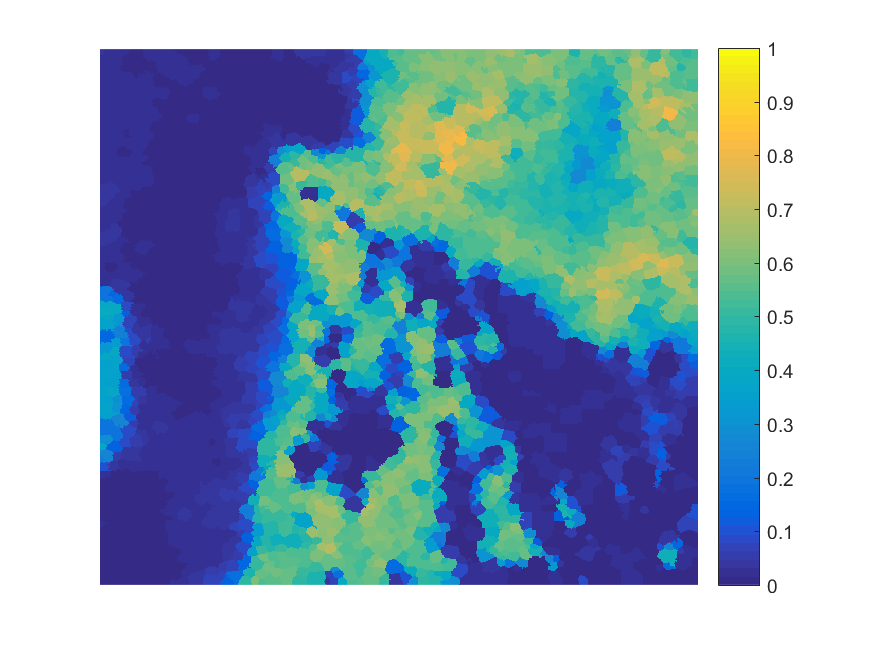}
		\label{fig:fhisthaups}
	}
	~
	\subfloat[PFLICM Cluster 2]{
		\includegraphics[scale=0.3]{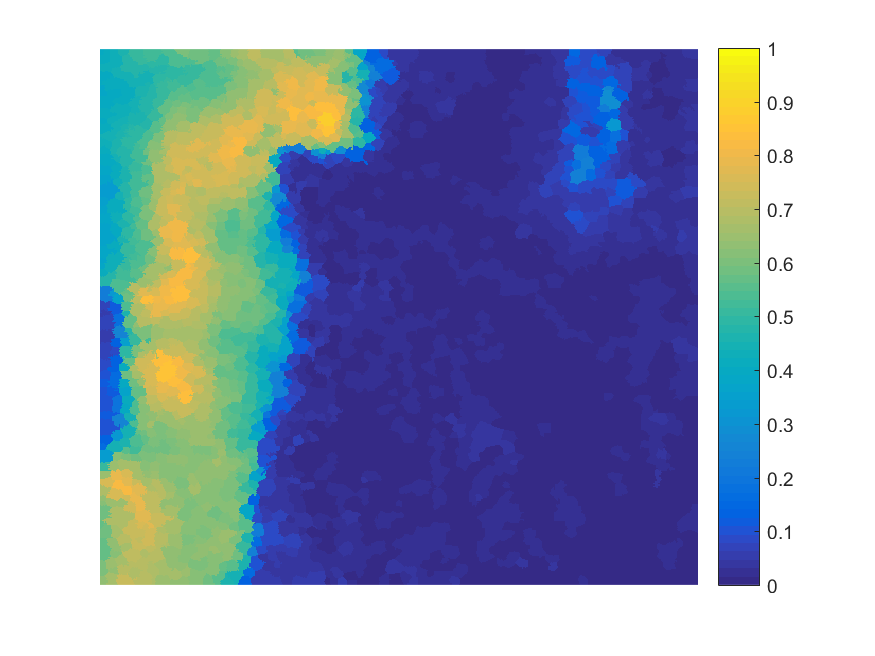}
		\label{fig:histmmps}
	}
	~
	\subfloat[PFLICM Cluster 3]{
		\includegraphics[scale=0.3]{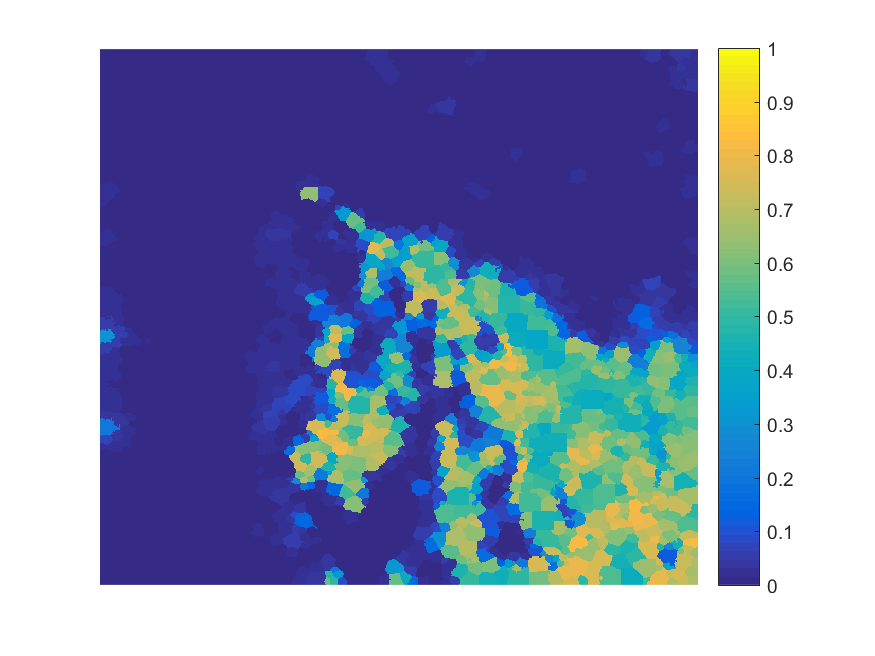}
		\label{fig:histminps}
	}
	
	\vskip 0pt
	\subfloat[PFCM Cluster 1]{
		\includegraphics[scale=0.3]{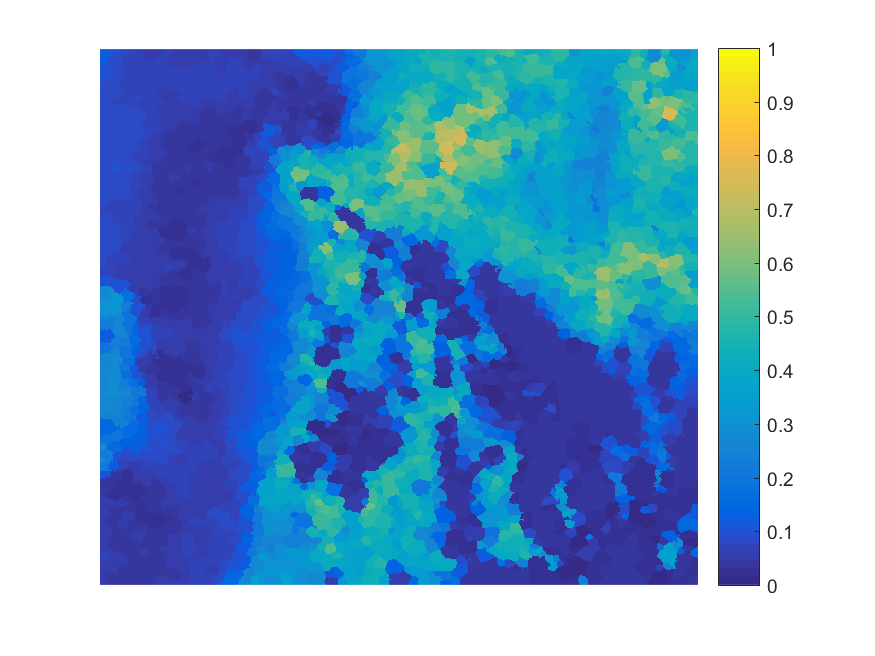}
		\label{fig:fhisthaups2}
	}
	~
	\subfloat[PFCM Cluster 2]{
		\includegraphics[scale=0.3]{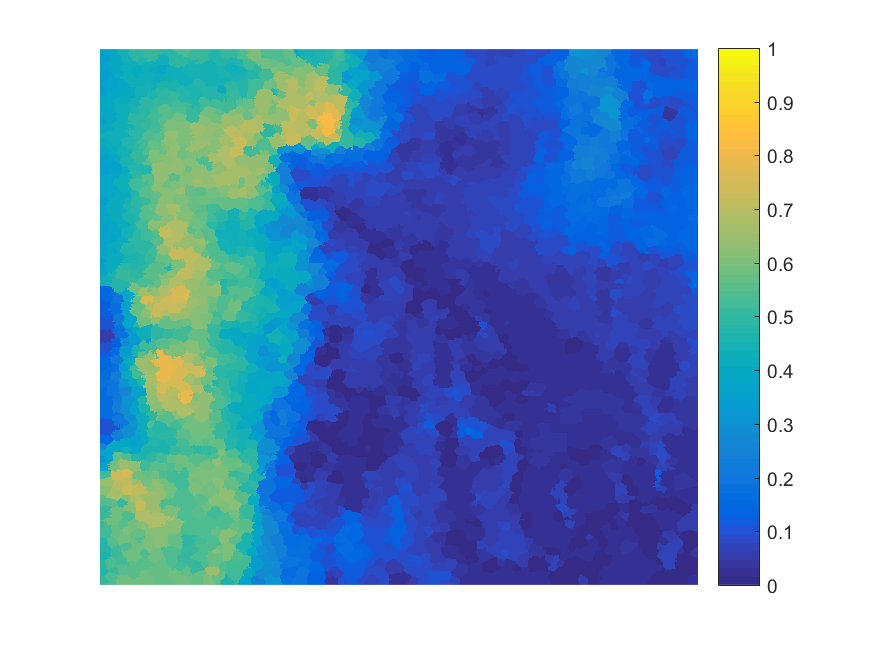}
		\label{fig:histmmps1}
	}
	~
	\subfloat[PFCM Cluster 3]{
		\includegraphics[scale=0.3]{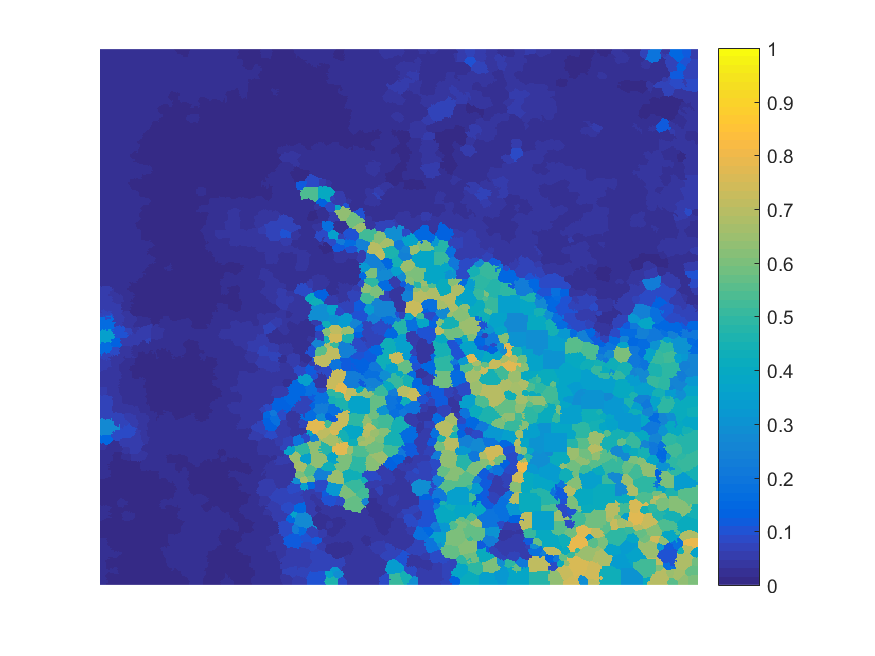}
		\label{fig:histminps2}
	}

	\caption{(a) SAS image with distinct regions: sand ripple, hard packed sand, and coral/rocky surface. The segmentation results for PFLICM (b-d) and PFCM (e-g) with selected features using VXB index.}
	\label{fig:mildps}
\end{figure}
\section{Discussion}
Cluster validity indices are usually used to determine the optimal number of clusters. The XB and VXB indices can also provide information on the best features to use to cluster the data. As shown in Figures 2 and 3, the segmentation results for the features that minimized the XB score are visually better than using all of the features. In comparing the two metrics, VXB is useful for algorithms that perform fuzzy and possiblistic clustering. The VXB accounts for typicality and as seen in Table 2, the PFLICM VXB index was the lowest score in comparison to all algorithms. The segmentation results agree with this quantitative measure. If only the XB index was used, K-Means clustering was the best algorithm. For the SAS imagery, transition regions will need to be accounted for so ``crisp" partitions will not be optimal for this application. For example, within this image, there are three seafloor textures. Between each seafloor texture, there is not a distinct boundary between where one seafloor type ends and the other begins. The K-Means algorithm clusters the sand ripple and hard packed sand regions together (Figure 3k)  while the other fuzzy clustering algorithms do not. The initial hypothesis for why the XB index might favor K-Means is that the membership of each pixel is either 0 or 1 leading to fewer terms in the calculation of the index as opposed to fuzzy clustering algorithms in which pixels can have some membership in each cluster. Each algorithm was implemented with the feature selected by K-means (Sobel) and the results of this test (averaged over 5 runs) are found in Table 5. N is the number of valid pixels in the image and C is the total number of clusters. Despite there being fewer non-zero terms in the compactness calculation (numerator of the XB index), each clustering algorithm has a compactness score on the same magnitude. The separation calculation (denominator of the XB index) is higher for K-Means which leads to a lower XB index. 
\begin{table}[ht]
	\caption{XB Index Calculation Details ($\pm$ 1 Standard Deviation)}  
	\label{tab:Paper Margins}
	\begin{center} 
		      
		\begin{tabular}{|c|c|c|c|c|} 
			\hline
			\rule[-1ex]{0pt}{3.5ex}   Algorithm & Numerator & Denominator & XB Score & Number of Non-zero Terms in Numerator\\
			\hline
			\rule[-1ex]{0pt}{3.5ex}    PFLICM & 2.009e4$\pm$0.000 &1.827e4$\pm$0.000 & 1.100$\pm$0.000 & N*C\\
			\hline
			\rule[-1ex]{0pt}{3.5ex}    PFCM & 1.749e4$\pm$0.000 & 1.038e4$\pm$0.000& 1.685$\pm$0.000 & N*C\\
			\hline
			\rule[-1ex]{0pt}{3.5ex}   FLICM  & \textbf{1.532e4$\pm$0.000} &1.894e4$\pm$0.000 & 0.809$\pm$0.000 & N*C\\
			\hline
			\rule[-1ex]{0pt}{3.5ex}    K-Means & 2.095e4$\pm$0.000 &\textbf{7.492e4$\pm$0.000}& \textbf{0.280$\pm$0.000} & N \\
			\hline
		\end{tabular}
	\end{center}
\end{table}
\\Figure 6 further validates the usefulness of the XB and VXB indices for feature selection. The randomly selected features (Figure 6a-c) chosen for segmentation were histogram of oriented gradients, local binary patterns and shape. The XB index selected lacunarity, Sobel, and homogeneity (Figure 6e-g) while the VXB index selected lacunarity, sobel and shape. The segmentation results for the selected features appear more distinct than those of the randomly chosen features. 

\section{CONCLUSION}
PFLICM with forward feature selection was presented in this work. This work has two main contributions: the use of cluster validity as evaluation metric for feature selection and a variation of the XB index for fuzzy and possiblistic clustering algorithms. Unsupervised feature selection is a more difficult than supervised applications since error cannot be used to determine how well a model performs with a subset of features. Quantitative measures such as compactness and separation of the clusters obtained from the features validated the qualitative results (segmentation maps). Several extensions of this work are possible such as using a different feature selection process (filter or embedded methods) and different evaluation metrics (other cluster validity measures\cite{liu2010understanding}, mutual information\cite{peng2005feature}, entropy).  

\acknowledgments 
 
This work was funded by the Office of Naval Research grant N00014-16-1-2323.  

\bibliographystyle{spiebib} 
\bibliography{SPIE_references} 

\end{document}